\documentclass{article}

\usepackage{microtype}
\usepackage{graphicx}
\usepackage{subcaption}
\usepackage{paralist}
\usepackage{booktabs} % for professional tables

\usepackage{hyperref}

\usepackage{todonotes}
\usepackage{caption}
\usepackage{helvet}  %Required
\usepackage{courier}  %Required
\usepackage{url}  %Required
\usepackage{graphicx}  %Required
\usepackage{multirow}
\usepackage{amsthm}
\usepackage{color}
\usepackage{MnSymbol}
\usepackage{makecell}
\usepackage{arydshln}
\usepackage{amsmath}
\usepackage{xcolor}
\usepackage{caption} 
\usepackage{natbib}
\usepackage{textcomp}
\usepackage{wrapfig}
\usepackage{algorithm}
\usepackage{algorithmic}
\usepackage{amsfonts}
\usepackage{multirow}

\newtheorem{theorem}{Theorem}[section]

\newtheorem{proposition}{Proposition}[section]

\newcommand{\W}{\mathbf{w}}
\newcommand{\U}{\mathbf{u}}
\newcommand{\X}{\mathbf{x}}
\newcommand{\V}{\mathbf{v}}
\usepackage[accepted]{icml2020}

\icmltitlerunning{Volumization as a Natural Generalization of Weight Decay}

\begin{document}

\twocolumn[
\icmltitle{Volumization as a Natural Generalization of Weight Decay}

% It is OKAY to include author information, even for blind
% submissions: the style file will automatically remove it for you
% unless you've provided the [accepted] option to the icml2019
% package.

% List of affiliations: The first argument should be a (short)
% identifier you will use later to specify author affiliations
% Academic affiliations should list Department, University, City, Region, Country
% Industry affiliations should list Company, City, Region, Country

% You can specify symbols, otherwise they are numbered in order.
% Ideally, you should not use this facility. Affiliations will be numbered
% in order of appearance and this is the preferred way.
\icmlsetsymbol{equal}{*}

\begin{icmlauthorlist}
\icmlauthor{Liu Ziyin}{equal,tk}
\icmlauthor{Zihao Wang}{equal,th}
\icmlauthor{Makoto Yamada}{ku}

\icmlauthor{Masahito Ueda}{tk}

\end{icmlauthorlist}

\icmlaffiliation{tk}{Department of Physics \& Institute for Physics of Intelligence, University of Tokyo}
\icmlaffiliation{th}{Department of Computer Science and Technology, Tsinghua University}
\icmlaffiliation{ku}{Kyoto University}

\icmlcorrespondingauthor{Liu Ziyin}{zliu@cat.phys.s.u-tokyo.ac.jp}
\icmlcorrespondingauthor{Zihao Wang}{wzh17@mails.tsinghua.edu.cn}

% You may provide any keywords that you
% find helpful for describing your paper; these are used to populate
% the "keywords" metadata in the PDF but will not be shown in the document
\icmlkeywords{Machine Learning, ICML}

\vskip 0.3in
]

% this must go after the closing bracket ] following \twocolumn[ ...

% This command actually creates the footnote in the first column
% listing the affiliations and the copyright notice.
% The command takes one argument, which is text to display at the start of the footnote.
% The \icmlEqualContribution command is standard text for equal contribution.
% Remove it (just {}) if you do not need this facility.

\printAffiliationsAndNotice{}  % leave blank if no need to mention equal contribution
%\printAffiliationsAndNotice{\icmlEqualContribution} % otherwise use the standard text.

\begin{abstract}
We propose a novel regularization method, called \textit{volumization}, for neural networks. Inspired by physics, we define a physical volume for the weight parameters in neural networks, and we show that this method is an effective way of regularizing neural networks. Intuitively, this method interpolates between an $L_2$ and $L_\infty$ regularization. Therefore, weight decay and weight clipping become special cases of the proposed algorithm. We prove, on a toy example, that the essence of this method is a regularization technique to control bias-variance tradeoff. The method is shown to do well in the categories where the standard weight decay method is shown to work well, including improving the generalization of networks and preventing memorization. %We thus propose the method, \textit{volumization}, as a natural generalization of weight decay. 
 Moreover, we show that the volumization might lead to a simple method for training a neural network whose weight is binary or ternary.
\end{abstract}

\vspace{-7mm}
\section{Introduction}

\begin{figure}[t]
    \centering
    \includegraphics[width=.8\linewidth]{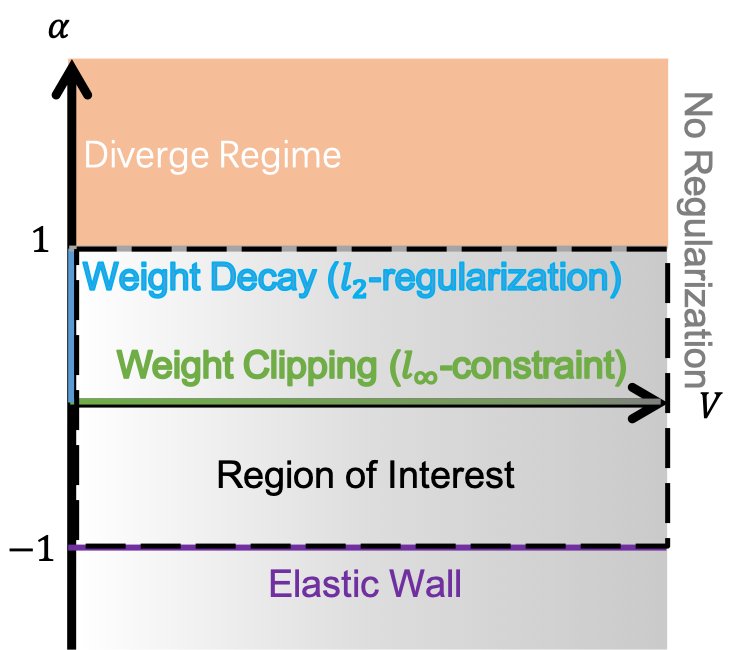}
    \vspace{-3mm}
     \caption{Diagram of various subclasses of \textit{volumization}$(V, \alpha)$, the proposed regularization method with hyperparamter $\alpha$ and $V$. When $\alpha>0,\ V=0$, volumization is equalent to weight decay with parameter $1-\alpha$. When $\alpha = 0,\ V>0$, the method is equivalent to performing a weight clipping at $V$ and $-V$. When $\alpha>0$ and $V>0$, one obtains an interpolation between weight decay and weight clipping. When $\alpha < 0$ and $v < 0$, weights are reflected at the virtual boundary $\pm V$.}\label{fig:phase-diagram}
    
\end{figure}
Regularization plays a central role in machine learning research. Deep learning systems, while hard to analyze analytically, also requires a combination of a series of regularization techniques to work well \cite{goodfellow2016deep, kukavcka2017regularization}. Popular regularization techniques in deep learning include but are not limited to weight decay \cite{krogh1992simple}, dropout \cite{srivastava2014dropout}, label smoothing \cite{muller2019does} etc.. One of the oldest yet the most popular approach is the weight decay, which is also related to $L_2$ regularization in the statistical learning literature. Weight decay has been shown to work well in improving the generalization and control overfitting of neural networks trained with stochastic gradient descent \cite{zhang2018three}, and applies equally well to other optimizers such as ADAM \cite{loshchilov2017decoupled}.

To motivate the proposed method, let us first review what we now know about neural networks. First of all, neural networks are often overparametrized and this overparametrization has been observed to be beneficial, in terms of the ease of training \cite{arora2018optimization, du2018gradient} and the genaralization ability \cite{geiger2019jamming, belkin2018reconciling}. Coming with the overparametrization is the difficulty to constrain and regularize a neural network. For example, it is shown that modern neural networks can easily memorize all the training data points when they are mislabeled, even when (1) data augmentation such as random crop is used, and (2) weight decay and dropout are applied \cite{Zhang_rethink}. One potential remedy to this problem is the recently discovered double-descent phenomenon \cite{belkin2018reconciling, nakkiran2019deep}, which shows that when the presence of label noise is mild (usually with corruption rate smaller than $20\%$), overparametrization itself can provide robustness to the model and no necessary measure needs to be taken. However, recent studies and experiments on label noise suggests that, when the noise rate is extreme, the neural network will still be negatively affected significantly by label noise \cite{ziyin2020learning, zhang2018generalized, hendrycks2019using}. This calls for the need of a stronger regularization method that can effectively limit the expressitivity of a neural network when needed, since such method does not exist yet. This paper aims at proposing a general regularization method with wider applicability than weight decay.

\textbf{Contribution:} The main contribution of this work is a generalized regularization method of weight decay, which (1) can improve where weight decay is traditionally shown to be effective, such as improving generalization of a model, (2) can be used to deal with problems where weight decay is ineffective, such as preventing memorization of overparameterized networks, and robustness to adversarial attacks, and (3) can even control the weight parameter distribution of neural networks, which is shown to have the potential to perform neural network binary quantization, even in the presence of extreme label noise.

%At the core of the mechanism for regularization methods is the universality of the bias-variance tradeoff, where an model cannot achieve the best generalization performance either because it does not converge to the optimal parameter, or because variance in the solution exists around its converged point. To be more formal, for many loss functions, the loss functions can be decomposed into two additive terms \cite{domingos2000unified}:
%\begin{equation}
%    \ell(\lambda) = \underbrace{b^2(\lambda)}_{\text{bias}}+ \underbrace{\sigma^2(\lambda)}_{\text{variance}},
%\end{equation}
%where $\lambda$ is the regularization parameter. 
%While raw methods (such as a maximum likelyhood estimator without regularization) often constitutes an unbiased estimator, its generalization ability might suffer from a high variance. 

%\begin{equation}
%    \lambda^* = \argmin_{\lambda} \ell(\lambda) = \argmin_{\lambda} b^2(\lambda) + \sigma^2(\lambda).
%\end{equation}
%When $\ell(\lambda)$ is differentiable, we might solve for $\lambda$ by taking derivative and set to zero to find $\lambda^*$
%\begin{equation}
%    \frac{\partial b^2(\lambda)}{\partial \lambda } = - \frac{\partial \sigma^2(\lambda)}{\partial \lambda }.
%\end{equation}

\vspace{-3mm}
\section{The Proposed Method}
\vspace{-1mm}
\begin{algorithm}[t]
	\caption{\textit{Volumization}$(V, \alpha)$}
	\label{alg:general volumization}
	\begin{algorithmic}
	\STATE
		\STATE {\bfseries Input:} weights at step $t$: $w_t \in \mathbb{R}^d$, its momentum $m_t \in \mathbb{R}^d$,  optimizer $G(w): \mathbb{R}^{2d} \to \mathbb{R}^{2d}$, volume parameter $V>0$, discount parameter $\alpha$
		\STATE $\hat{w}_{t+1}, \hat{m}_{t+1} = G(w_{t}, m_t)$ // optimizer step
		\IF{$|\hat{w}_{i, t+1}| > V$}
		\STATE \quad\quad $w_{i, t+1}= \hat{w}_{i, t+1} + (1-\alpha)(V \times \texttt{sgn}(\hat{w}_{i, t+1}) -\hat{w}_{i, t+1})$
		\STATE \quad\quad $\hat{m}_{i, t+1} = \alpha \times \hat{m}_{i, t+1}$
		\ENDIF
		%\ENDFOR
	\end{algorithmic}
\end{algorithm}

\begin{figure}
    \centering
    \includegraphics[width=\linewidth]{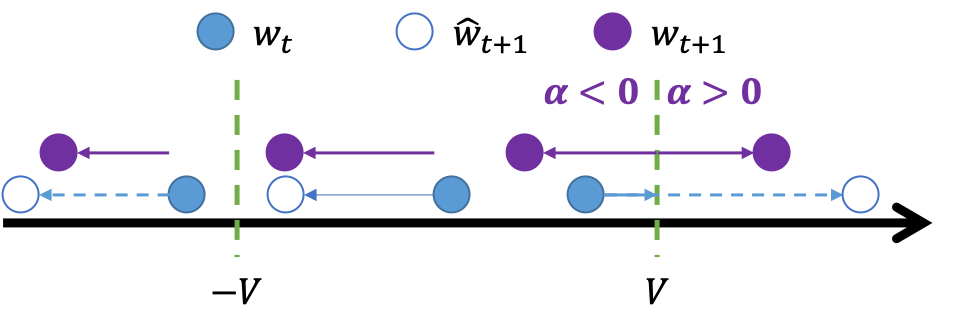}
    \vspace{-8mm}
    \caption{
    Illustration of volumization. Particles indicate the weight and arrows indicate the update step. Dashed arrows are the parts out of the interval $[-V, V]$ and will be volumized. When $\alpha < 0$, volumization behaves like reflection on the wall}
    \label{fig:motivation}
\end{figure}

One recently recognized way to effectively constrain the expressivity of modern neural networks is to constrain the largest eigenvalue of the weight matrices, since this directly limits the Lipschitz constant of the network, which is an important measure of the expressivity and the robustness of the network \cite{virmaux2018lipschitz, gouk2018regularisation}, and one straightforward way to achieve this is to contrain the magnitude of every element in a weight matrix. This is the basis for volumization. Formally, we propose to confine all the weights within an interval $[-V, V]$ defined by the volume parameter $V$. However, constraining the each parameter directly might be an overkill when very strong regularization is not needed, we introduce a ``softening" parameter $\alpha$ to introduce soft boundaries at $V$ by decaying parameters towards $V$ at rate $1-\alpha$, we call $V$ the ``volume", and $\alpha$ ``weight decay". 

%Let's consider the SGD weight updating at $t+1$ step as an example, $\hat{w}_{t+1} = w_t - \lambda m_t $. We make such an analogy between optimizer update and 1-D particle movement: learning rate $\lambda$ is the ``time'', $-m_t$  is the ``velocity'', and each gradient descent step is corresponding to ``particle movement with constant speed''.\todo{more professional description}
%Volumization is the regularization process for updated weight $w_{t+1}$ from $\hat{w}_{t+1}$.
%Three situations need to be discussed:

%\begin{compactitem}
%    \item[(1)] $\hat{w}_{t+1} \in [-V, V]$: we set $w_{t+1} = \hat{w}_{t+1}$ directly
%    \item[(2)] $\hat{w}_{t+1} \notin [-V, V], w_t \in [-V, V]$: we discount the momentum when the weight particle is out of the volume interval. As shown in Figure~\ref{fig:motivation} (b), we have volumized update rule $w_{t+1} = \hat{w}_{t+1} + (1-\alpha)(V - \hat{w}_{t+1})$;
%    \item[(3)] $\hat{w}_{t+1} \notin [-V, V], w_t \notin [-V, V]$:  compared to case (2) that particle crossing the boundary, there is an additional positive term $(w_t-V) (1-\alpha)$ as shown in Figure~\ref{fig:motivation} (a). We ignore this term for simplicity so that the volumized update rule is the same as situation (2). This update rule actually impose stronger regularization for weights out of the volume.
%\end{compactitem}

The algorithms is given in Algo.~\ref{alg:general volumization}. Notice that $\alpha$ can be both positive and negative. When $\alpha > 0$, the constraint is soft; when $\alpha\leq 0$, the constraint is hard. While the optimal solution for all $\alpha\leq 0$ is the same,  we note $\alpha$ affects the learning dynamics, and the learned model will potentially converge to different solutions, in this work, we focuses on studying the case when $\alpha \geq 0$, leaving the negative case to future work. We note that this method may be plugged into any commonly used optimizers, which can be defined as a mapping function that takes a weight $w_t$ at time step $t$ and maps to a weight parameter at time step $t+1$. We give examples on volumized SGD and volumized ADAM in the appendix.%\todo{TBA}.

A summary of the volumization in the $V-\alpha$ plane is given in Figure~\ref{fig:phase-diagram}, connecting volumization with other regularization%. When $\alpha=1$ or $V$ is large enough, volumization practically adds no regularization to the original optimizer. When $\alpha >1$, the momentum for the weight outside the volume will be exponentially amplified, so the parameter will diverge during training. %For $\alpha < -1$, the momentum may not be amplified at all steps, but the weight may also diverge.
We note that three parameter settings appear especially interesting:

\begin{description}
    \item[Elastic Wall $(V>0,\  \alpha=-1)$] This is equivalent to having a physical wall at $\pm V$, and, treating the parameters as particles, the collision between parameters and the walls are elastic, with no energy loss. By standard assumptions in statistical physics, the final solution converges to the Boltzmann distribution $\sim \exp[ - \ell(w)]$ in the volume $[-V, V]$ \cite{zhang2018theory};
    \item[Weight Clipping  $(V>0,\ \alpha=0)$] This is literally performing a parameter clipping at every optimization step at $\pm V$. In physical terms, this is equivalent to have a completely inelastic collision between the wall and the parameters, after which all the kinetic energy of the parameters are lost, and this causes parameters to concentrate at $\pm V$;
    \item[Weight Decay $(V=0,\ 0 < \alpha < 1)$] This is equivalent to \textit{weight decay}, where $1-\alpha$ is equivalent to the standard weight decay parameter. The only difference is that we also decay the momentum by $1-\alpha$, but this is often negligible since in practice the weight decay strength is at the order of $10^{-4}$.
\end{description}
%In this paper, we focused on the zone $(V,\ \alpha)\in[0, +\infty] \times [-1, 1]$.

\begin{figure}
    \centering
    \vspace{-2mm}
    \includegraphics[trim=40 0 0 0, clip,width=0.8\linewidth]{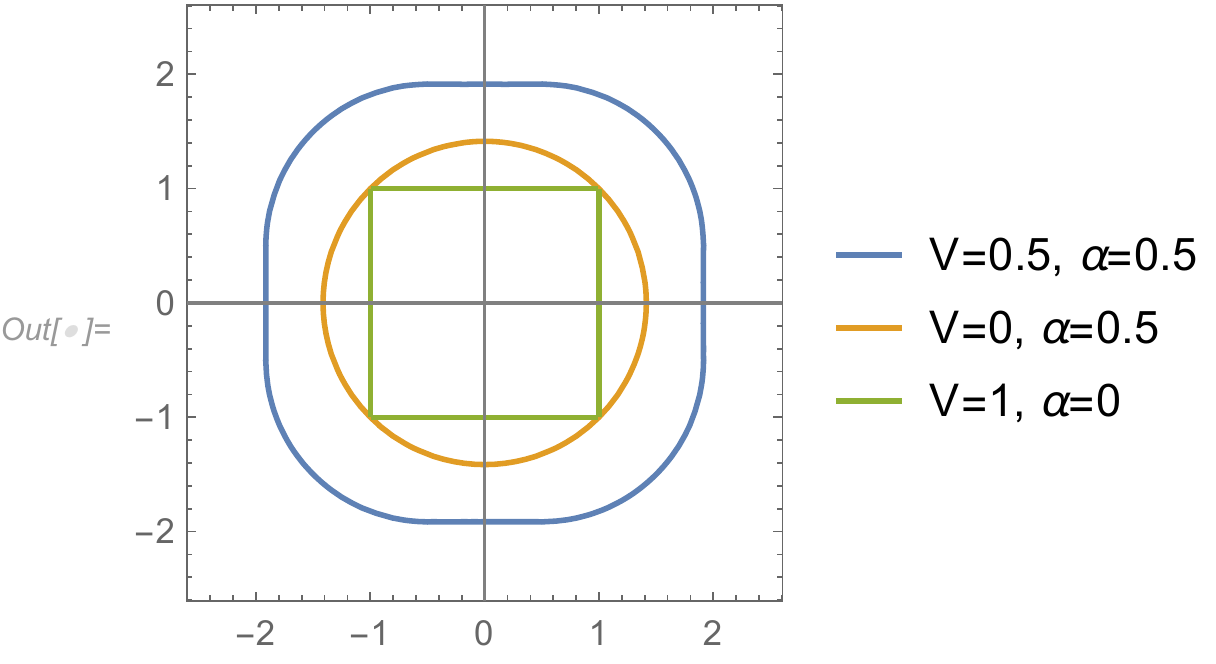}
    \vspace{-5mm}
    \caption{Geometry of volumization. Plotted are the equi-potential lines for three different parameter setting. One sees that, when $V=0$, the geometry is $L_2$; when $V=1$, the geometry is $L_\infty$, and with $V>0, \alpha >0$, the methods interpolates between the two.}
    \label{fig:geometry}
    \vspace{-5mm}
\end{figure}

An alternative way to understand volumization is that it is an interpolation between imposing a $L_\infty$ and $L_2$ geometry on the model parameters. See Figure~\ref{fig:geometry}. Notice that one potential problem with this method is that different weight matrices might require a different $V$, and so the number of hyperparameters might grow with the depth of the network, but this is not the case, since there exists a very natural definition of $V$ for each layer. In particular, it is shown that there is a critical point $a$ such that one should initialize the weights of neural networks uniformly in the region $[-a,a]$, where $a= \sqrt{6/h}$, where $h$ is the width of the network (or the number of fan-out channels) \cite{he2015delving}. $\sqrt{6/h}$ is the critical initialization in the sense that, when $a\neq \sqrt{6/h}$, the variance of the activations will either explode or vanish. More importantly, initializing at this critical $a$ not only improves optimization of networks, but also improves the generalization of the learned model \cite{he2016deep}. This suggests that, for each layer, $V$ should be naturally defined with reference to $a$. In this work, we parametrize $V$ by $V=va$.
For example, setting $v=0.8$ means that we set the volume of every layer to be $0.8$ times $\sqrt{6/h}$. In the experiment section, we demonstrate that this indeed achieves good result.

\vspace{-3mm}
\section{Volumization Can Improve Generalization}
\vspace{-1mm}
In this section, we study a linear perceptron and show that volumization can improve generalization in a noisy setting. The mechanism of $V=0,\ \alpha>0$ is equivalent to weight decay and has been intensively studied by previous works \cite{krogh1992generalization, hastie2019surprises} and so we focus on the cases when $V>0,\ \alpha=0$. We believe that the case when both $V,\ \alpha \neq 0$ can be even more helpful to combating noise, but we leave this to be demonstrated experimentally.
\vspace{-3mm}
\subsection{Warmup: General Framework and Preliminaries}
\vspace{-1mm}
In this section, we set the notation and introduce the problem by analyzing a regularization-less setting. As in a series of previous works that have been shown very successful \cite{gardner1989three, krogh1992generalization, hastie2019surprises, liang2020think}, we assume a teacher-student setting, where we want to train a network $f_w$, with weight $w$, on a task whose target is produced by a teacher network $f_u$. Let $\{x_i\}_{i=1,...,N}$ denote the dataset, and $f_u(x_i)$ their corresponding labels. We define the \textit{empirical generalization error} to be:
\begin{equation}
    \ell(w) = \frac{1}{2N} \sum_i \left[ f_w(x_i) - f_u(x_i) \right]^2 
\end{equation}
The generation error is defined as the expected value of this over all samplings of dataset. I.i.d. noises are present in the training set, such that the effect is that the labels are shifted by a noise with variance $\sigma^2$: $f_u(x_i) \to f_u(x_i) + \epsilon_i$. Denoting the difference $f_w(x_i) - f_u(x_i)$ as $b_i$, the training objective is then
\begin{align}\label{eq: noise}
    \Tilde{\ell}(w) &= \frac{1}{2N}\sum_i \left[ f_w(x_i) - f_u(x_i) - \epsilon_i \right]^2  %\\
    %&= \sum_i  b_i^2 + \epsilon_i b_i  + \epsilon_i^2
\end{align}

Now we make a simplification to allow for analytical analysis of how our method works in the presence of label noise. We consider a case where both $f_u$ and $f_w$ are linear networks parametrized by a linear transformation $\mathbf{w}\in \mathbb{R}^d$ and $\mathbf{u}\in \mathbb{R}^d$ respectively, and we use non-bold letter with subcript to denote each of the elements of these vectors, for example, $w_i$ denotes the $i$-th element of $\W$. Plugging in, 
\begin{equation}
    \ell(\mathbf{w}) = \frac{1}{2N} \sum_i ( \W^T\X_i -\U^T\X_i )^2 
\end{equation}
the update rule of gradient descent (GD) is then
%\begin{align}
%    \dot{\W} &= - \frac{1}{N} \sum_i \left( \W^T\X_i -\U^T\X_i\right) \X_{i}\\
%    &= - \frac{1}{N}  \sum_k (w_{k} - u_{k}) \sum_i x_{ik} x_{ij}
%\end{align}

\begin{align}
    \dot{\W}^T &= - \frac{1}{N} \sum_i \left( \W^T\X_i -\U^T\X_i\right) \X_{i}%\\
    %&= - \frac{1}{N}  \left( \W^T -\U^T\right) \sum_i \X_{i} \X_{i}^T
\end{align}
Notice that we are taking the continuous time limit of GD (also called gradient flow), and this has been shown to give accurate approximation to the finite step GD algorithm \cite{mei2019mean}. Define the residual $\V:= \W - \U$, and the correlation matrix $A = \sum_i \X_{i} \X_{i}^T$ we obtain:
\begin{align}
    \dot{\V}^T &= - \frac{1}{N}  \V^T A;
\end{align}
when the data correlation matrix $A$ is full rank, we know that $\V\to 0$ with an exponential time scale $\sim e^{-\lambda t}$ where $\lambda$ is the smallest eigenvalue of $A$. As $t\to \infty$, we see that $\ell\to 0$, achieving perfect generalization.

When the noise is present as in Equation~\ref{eq: noise}, we may proceed as before to obtain the update rule
\begin{align}
    \dot{\V}^T &= - \frac{1}{N}  \V^T A -\frac{1}{N} \sum_i \epsilon_i \X^T_{i}
\end{align}
We diagonalize $A$ to obtain (the change in index $i\to r$ refers to this change in basis):
\begin{equation}
    \dot{v}_{r} = - \frac{1}{N}  A_r v_r -\frac{1}{N} \sum_i \epsilon_i x_{ir}
\end{equation}
where $r$ denotes the diagonalized indices, and the solution at $t \to \infty$ is 
\begin{equation}
    v_r = \frac{- \frac{1}{N} \sum_i \epsilon_i x_{ir} }{A_r}
\end{equation}
Thus the effect of noise can be modeled by a shift in the parameters by a random variable proportional to $\epsilon$. 
The generalization error is then
\begin{align}
    \mathbb{E}_{\X, \epsilon_i} \left[\sum_r^d v_r^2\right]  %\frac{\frac{1}{N^2} \mathbb{E} [(\sum_i \epsilon_i x_{ir})^2] }{A_r^2}\\
    %&= \frac{\frac{1}{N^2}  \sum_i \mathbb{E}[\epsilon_i^2] \mathbb{E}[x_{ir}]^2 }{A_r^2}\\
    &= d\frac{  A_r^2\sigma^2 }{N A_r^2} + o\left(\frac{d}{N}\right) = \frac{d\sigma^2}{N} + o\left(\frac{d}{N}\right),
\end{align}
i.e., the generalization error is proportional to the strength of the noise. This is a standard result that was also obtained in other works using similar or different methods \cite{krogh1992simple, hastie2019surprises}.
\vspace{-3mm}
\subsection{Volumization in the Presence of Noise}\label{Sec: volumization theory example}
\begin{figure}[t]
%\centering
%\caption{Quantization of an LSTM network trained on the IMDB dataset.}\label{fig: quantize}
\begin{subfigure}[b]{0.55\linewidth}

    \centering
    \vspace{-3mm}
    \includegraphics[trim=0 0 0 8, clip,width=\linewidth]{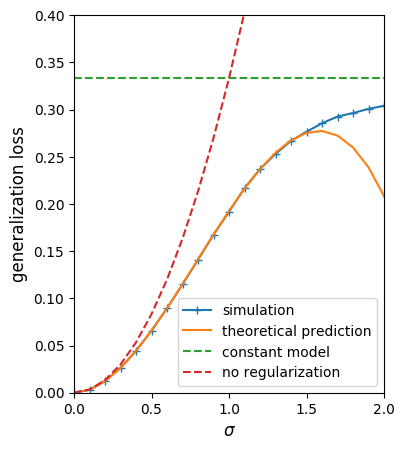}
    \vspace{-8mm}
    \caption{Uniform Noise}

\end{subfigure}
\hfill
%\centering
\begin{subfigure}[b]{0.4\linewidth}
    \centering
    \vspace{-3mm}
    \includegraphics[trim=0 0 0 8, clip,width=1\textwidth]{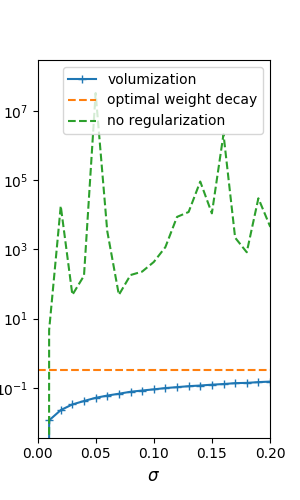}
    \vspace{-8mm}
    \caption{Cauchy Noise}
\end{subfigure}
\vspace{-4.5mm}
\caption{(a) Comparison of the prediction in Theorem~\ref{theo: optimal} with simulation for $a=1$. \textit{Constant model} refers to a model with all parameters being $0$. While our theory assumes $\sigma<1$, the calculated functional form agrees with experiment up to $\sigma \approx 1.5$. (b) Experiment when the noise obeys a $\sigma$-scaled standard Cauchy distribution. This shows the effectiveness of volumization when the noise is extreme. Here no regularization has a divergent error, while optimal weight decay converges to a constant model; volumization, however, succeeds in learning this task.}\label{fig:theo uniform}
\vspace{-4mm}
\end{figure}

\vspace{-1.5mm}
To proceed, we make the assumption that the effect of label is to shift the target variables $u_k' = u_k + \eta_k$. In this setting, we show that volumization with reasonable $V$ can always reduce generalization error at the optimal minimum, denoted by $b^2(V)$, compared to without volumization (i.e. when $V\to \infty$) at optimal solution, denoted by $b^2$. Formally, we prove:
\begin{theorem}
    Let the teacher weights $u_k$ follow a uniform distribution $Unif(-a, a)$, and let the noises $\eta_k$ follow a uniform distribution $Unif(-\sigma, \sigma)$, 
    and the $w_k$ be the learned parameters using volumization algorithm with $V\geq 0$ and $\alpha=0$, then
    \begin{equation}
        \mathbb{E}_{\mathbf{\eta}}\left[\frac{1}{d}||\W - \U||_2^2 \right] \geq \bigg(1 - \frac{27\sigma}{64a}\bigg)\frac{\sigma^2}{3}.
    \end{equation}
    with equality achieved at $V^*=a-\sigma/2$. Moreover, when the correlation matrix $A= I$, the the above equation is also equal to the generalization error, and with optimal $V$ being $V^*$.
    %then $b^2(V) \leq b^2$ when $V \in (a-\sigma, a + \sigma)$, where the equality is only achieved when either $\sigma=0$ or $\sigma=a$.
\end{theorem}
The proof is given in the appendix, and $A$ can be easily made diagonal by preprocessing the dataset by the inverse correlation matrix. One curious point is that the maximum gain is obtained when the signal-to-noise ratio (SNR) is around $1$.  One can also show that the generalizatoin error can be reduced for a range of $V$:
\begin{theorem}\label{theo: optimal}
    Let the problem be defined as in the previous theorem, the optimal solution exists at $V= a- \frac{\sigma}{2}$, then $b^2(V) \leq b^2$ for nay $V \in (a-\sigma, a + \sigma)$, where the equality is only achieved when either $\sigma=0$ or $\sigma=a$.
\end{theorem}
Intuitively, this proof is based on the idea using volumization induces a bias in the model parameters, while controlling the model variance, thus leading to a control over the bias-variance tradeoff. The theoretical prediction in compared to simulation in Figure~\ref{fig:theo uniform}. The simulation is obtained by searching for grid-searching for the optimal $V^*$ on the task described in the experiment section. We see that in the region $\sigma\in (0, 1.5)$, the theory gives perfect agreement with simulation (note that in the derivation of the theory, we assumed  $\sigma\leq 1$) To be complete, we also cite the the result for the when $V=0,\ \alpha>0$ (i.e., for weight decay):
\begin{theorem}
    \cite{krogh1992simple} On the same problem, and let $A=I$, the generalization error of weight decay is $\frac{\sigma^2 a^2}{3({\sigma^2}+ a^2)}$ per parameter, at optimal weight decay parameter $1 - \alpha = \frac{\sigma^2}{a^2}$.
\end{theorem}
While this is of similar magnitude to volumization for the example we studied, it converges to a constant model with all parameters being $0$ when the noise has unbounded variance. To illustrate this, we plot this (optimal weight decay) versus simulated volumization in Figure~\ref{fig:theo uniform} when the noise obeys a $\sigma$-scaled standard Cauchy distribution, we see that volumization with $V>0,\ \alpha=0$ is the only method to learn such a problem. This shows that volumization can help with learning when the noise is extreme. For deep, non-linear networks, the analysis is in general hard; we resort to demonstrating the usefulness of volumization through extensive experiments in this case.

%One can also compare this result on the same problem with the %generalization error of weight decay:
%\begin{theorem}
%    (\cite{krogh1992simple}) On the same problem, the generalization error of weight decay at optimal weight decay parameter $\lambda = \frac{\sigma^2}{a^2}$ is $\frac{\frac{4}{3}\sigma^2 a^4}{(\sigma^2 + a^2)^2}$.
%\end{theorem}

%\begin{figure*}[t]
%    \centering
%    \includegraphics[width=\linewidth]{plots/mnist-v=0_5.png}
%    \includegraphics[width=\linewidth]{plots/mnist-alpha=-0_99.png}
%    \vspace{-4mm}
%    \vspace{-4mm}
%    \caption{Test performance on the MNIST dataset given $v=0.5$ or $\alpha=-0.99$ under varying label noise conditions}
%    \label{fig:mnist-hpeff}
%\end{figure*}  

\vspace{-2mm}
\subsubsection{Controlling Weight Distribution}
\vspace{-1mm}
One additional effect results from the above analysis. Training with volumization with volume $V$ results in a parameter distribution with two peaks located at $V$ and $-V$, and these peaks attracts more density as we reduce $V$. This can be used to control the distribution of weight parameters of neural networks. Volumization with small or zero $\alpha$ naturally results a softly binarized network (binarized to $-V$ and $V$), and during inference, one might also move the remaining parameters also to $V$ and $-V$ to achieve an exact binarization of neural networks. We study this part in Section~\ref{exp: quantization}.

\vspace{-3mm}
\section{Some Useful Properties}
\vspace{-1.5mm}
One key property of the method when training with $\alpha \geq 0$ is that each element in the weight matrices are bounded from above and below by $\pm V$. A consequence is that the eigenvalue of the learned matrix is bounded.
\begin{proposition}
    Let $V\geq 0$, and a weight matrix $W\in \mathbb{R}^{a\times b}$ have elements $-V\leq W_{ij}\leq V$, and let $s_{max}(W) $ denote the largest singular value of $W$, then $ s_{max}(W)\leq V\max(a,b)$.
\end{proposition}
This prompts us to choose $V$ as a function of the size of weight matrix so as to control the largest eigenvalue, and motivates for a different $V$ value for each layer. Moreover, this fact can be used to bound the Lipshitz constant of a trained neural network.
\begin{proposition}
    let $d$ be the depth of a network, and define a feedforward neural network $f(\cdot)$ by 
    $f(x) = W_1 \sigma W_2 \sigma  ... \sigma W_d x $, where $\sigma$ is a $1-Lipshitz$ activation function, and each $W_i \in \mathbb{R}^{a_i\times b_i}$ is trained with volumization parameter $V_i = 1/\max(a_i,b_i)$ and $\alpha \geq 0$, then $f(\cdot)$ is a $1-Lipshitz$ function.
\end{proposition}
While the proof of these two results are elementary, they can be of great practical and theoretical use. One famous example is the training of a Wasserstein GAN / autoencoer using the Kantorovich dual, where the optimization domain is the set of all $1-Lipshitz$ functions \cite{arjovsky2017wasserstein}.

\vspace{-3mm}
\section{Experiments}
\vspace{-1mm}
In this section, we present experimental results on showing the usefulness of volimization in (1) improving the generalization performance of modern large-scale model on large image classification dataset; (2) improving robustness of the trained models to label noise and adversarial attacks and even achieving the state-of-the-art results, especially when the noise level is extreme; (3) controlling the weight distribution of neural networks, which can be directly used to train binarized and ternarized networks.
%(1) We present the direct experimental results supporting that volumization can improve generalization by training state-of-the-art models on large scale image classification tasks.
%(2) we conduct extensive experiments to reveal how volumization controls the bias-variance trade-off on varying label noise condition and different types of neural networks.
    %Our discussion includes fully-connected Dense Neural Network (DNN), Convolutional Neural Network (CNN), Recurrent Neural Network (RNN) as well as ATTention network (ATT).
    %We decouple the weight initialization and volumization by consider volume $V=vM_l$, where $M_l$ is the maximum absolute value by the initialization of the $l$-th layer.

The effect of volumization with hyperparameters is also discussed, we denote the volumization case as $\textrm{Vol}(v,\ \alpha)$.
%(3) we further demonstrate that neural networks could be more Lipshitz continuous when trained by neural networks.
Unless otherwise specified, we take ADAM as the default optimizer, the default learning rate is $10^{-4}$ and the batch size is 128.

\vspace{-3mm}
\subsection{Improving Generalization}
\vspace{-1mm}
\begin{table}
\centering
\caption{Top1/5 Accuracy of Fine-tuned EfficientNets on Tiny ImageNet; The column ``Best'' indicates the best test accuracy score and the colume Last indicates the average scores of last 10 epochs.}\label{table:efficientnet}
{\scriptsize
\begin{tabular}{llllllll}
\hline
\multirow{2}{*}{Model} & \multirow{2}{*}{Regularization} & \multicolumn{3}{c}{Top1 Accuracy} & \multicolumn{3}{c}{Top5 Accuracy} \\ \cline{3-8} 
    & & Best & AVE & Gap & Best & Last & Gap \\ \hline
\multirow{2}{*}{b0}
    & WD $5\times 10^{-4}$ & 78.16 & 76.76 & 1.40 & \textbf{93.52} & 92.71 & 0.81 \\
    & $\textrm{Vol}(1, 0.99)$ & \textbf{78.87} & \textbf{78.39} & 0.48 & 93.47 & \textbf{93.24} & 0.23\\
\hline
\multirow{2}{*}{b1} 
    & WD $5\times 10^{-4}$ & 79.93 & 75.65 & 4.28 &94.41 & 92.17 & 2.24\\
    & $\textrm{Vol}(1, 0.99)$ & \textbf{80.19} & \textbf{79.61} & 0.58 & \textbf{94.45} & \textbf{93.55}& 0.90 \\ 
\hline
\multirow{2}{*}{b2} 
    & WD $5\times 10^{-4}$ & 79.89 & 78.26 &1.63& 94.71 & 93.33 &1.38\\ 
    & $\textrm{Vol}(1, 0.99)$ & \textbf{81.90} & \textbf{81.27} &0.63& \textbf{95.02} & \textbf{94.34} &0.68\\
\hline
\multirow{2}{*}{b3} 
    & WD $5\times 10^{-4}$ & 83.21 & 81.34 &1.87& \textbf{95.54} & 94.59& 0.95\\ 
    & $\textrm{Vol}(1, 0.99)$ & \textbf{83.62} & \textbf{82.76} &0.86& 95.49 & \textbf{94.92}&0.57 \\
\hline
\end{tabular}}
\vspace{-1.5em}
\end{table}

We demonstrate that volumization can improve generalization by finetuning EfficientNets (from b0 to b3)~\cite{tan2019efficientnet}, the state-of-the-art image classification network. We initialize the network by the pre-trained weights from ImageNet.
We train EfficientNet-b\{0-3\} on the released Tiny ImageNet~\cite{wu2017tiny} train and validation set, respectively.
Tiny ImageNet has 200 classes. 
Each class has 500 training images, 50 validation images.
We grid-search the weight decay strength in the commonly used range $5\times 10^{-4}$ from $\{5\times 10^{-3}, 5\times 10^{-4}, 5\times 10^{-5}\}$ by cross validation in EfficientNet-b0 network.
For volumization, we pick volumization parameter $(v,\ \alpha) = (1, 0.99)$. The hyperparameters are kept the same for all EfficientNet-b\{0-3\} for simplicity, showing volumization does not need heavy parameter-tuning to work.
%By doing this, the test accuracy trajectory for each case reaches the highest accuracy in the first 10 epochs, and then decays at the last 10 epochs.
We train the network for 20 epochs by ADAM with weight decay or volumization and report the best value (Best) and last 10 epoch average (Last) of the test accuracy trajectory.
%We note that Best and Last value describes briefly how complex model overfits the benchmark dataset.
We can see from the table~\ref{table:efficientnet} that for all EfficientNet-b\{0-3\}, volumization achieves better top 1 Best and Last accuracy than weight decay.
For top 5 results, volumization is compatible to weight decay at Best accuracy but has better performance at Last accuracy.
Moreover, the gap between Best and Last values indicates the overfitting. Therefore, smaller gap indicates less overfitting and better generalization. Specifically, when model is trained by volumization, it produces consistently smaller gaps between the Best and Last value than weight decay.
%This result supports our claim that volumization improves generalization better than weight decay.

\vspace{-3mm}
\subsection{Robustness to Label Noise}% and $(v,\ \alpha)$ effect}
\vspace{-1mm}

\begin{table}[t]
\centering
\caption{Detailed comparison between weight decay and volumization under varying label noise conditions.$\textrm{Vol}_{\textrm{best}}$ and $\textrm{Vol}_{\textrm{Last}}$ are selected volumized cases by its best test accuracy and the average test accuracy of last 10 epochs, respectively.}\label{table:label-noise}
{\scriptsize\begin{tabular}{lllll}
\hline
\multirow{2}{*}{Model@Dataset} & Noise & \multirow{2}{*}{Regularization} & \multicolumn{2}{c}{Accuracy} \\\cline{4-5}
 & Ratio & & Best & Last \\\hline
\multirow{10}{*}{DNN @ MNIST} 
 & \multirow{5}{*}{0.4}
   & No Reg. & 96.50 & 89.65\\
 & & WD $5\times 10^{-3}$ & 94.82 & 94.70\\
 & & WD $5\times 10^{-4}$ & \textbf{96.23} & 95.92\\
 & & WD $5\times 10^{-5}$ & 96.10 & 95.42 \\
% & & $L_{q}$ & 96.37 & 95.78 \\
 & & Vol(1, 0.9999) & 96.18 & 95.40\\\cline{2-5}
 & \multirow{5}{*}{0.8}
   & No Reg. & 87.04 & 35.88\\
 & & WD $5\times 10^{-5}$ & 87.29 & 53.67\\
 & & WD $5\times 10^{-4}$ & 86.51 & 61.26\\
 & & WD $5\times 10^{-3}$ & 88.04 & 86.87\\
 & & Vol(1, 0.99) & \textbf{87.96} & 86.57 \\
\hline
\multirow{10}{*}{ResNet18 @ CIFAR10}
 & \multirow{5}{*}{0.4} 
   & No Reg. & 78.16 & 54.84\\
 & & WD $5\times 10^{-5}$ & 78.43 & 54.48\\
 & & WD $5\times 10^{-4}$ & 78.32 & 54.20\\
 & & WD $5\times 10^{-3}$ & 79.46 & 67.12\\
% & & $L_{q}$ & \textbf{83.29} & \textbf{82.34} \\
 & & Vol(0.25, 0.5) & \textbf{81.57} & \textbf{72.11}\\
 \cline{2-5}
 & \multirow{5}{*}{0.8}
   & No Reg. & 39.67 & 16.67\\
 & & WD $5\times 10^{-5}$ & 36.91 & 16.30\\
 & & WD $5\times 10^{-4}$ & 37.95 & 16.47\\
 & & WD $5\times 10^{-3}$ & 27.69 & 18.59\\
 & & Vol(0.25, 0.5) & \textbf{45.56} & \textbf{38.16} \\
\hline
\multirow{10}{*}{LSTMATT @ IMDB} & \multirow{3}{*}{0.2}
 & No Reg. & 68.56 & 64.26\\
 & & WD $5\times 10^{-5}$ & 73.43 & \textbf{68.08}\\
 & & WD $5\times 10^{-4}$ & \textbf{74.37} & 67.71\\
 & & WD $5\times 10^{-3}$ & 69.47 & 66.30\\
 & & Vol(0.25, 0.5) & 73.58 & 65.65 \\
 \cline{2-5}
 & \multirow{5}{*}{0.4} 
 & No Reg. & 57.18 & 55.30\\
 & & WD $5\times 10^{-5}$ & 57.93 & 54.37\\
 & & WD $5\times 10^{-4}$ & 60.99 & 55.38\\
 & & WD $5\times 10^{-3}$ & 55.18 & 50.00\\
 & & Vol(0.25, 0.5) & \textbf{65.02} & \textbf{55.53} \\
 \hline
\end{tabular}}
\vspace{-2em}
\end{table}
\begin{table}
\centering
\caption{Comparison with the state-of-the-art label noise method and the quantization experiments at extreme noise rate. We see that the proposed method achieves the state-of-the-art performance at this noise rate. What is surprising here is that quantization seems to make the model even more robust to lanbel noise.}\label{tab: benchmark}
{\footnotesize
\begin{tabular}{llll}
\hline
\multirow{2}{*}{Model @ Dataset - Noise Ratio} & \multirow{2}{*}{Method} & \multicolumn{2}{c}{Accuracy} \\\cline{3-4}
 & & Best & Last \\\hline
% \multirow{2}{*}{DNN @ MNIST - 0.8}
%  %& NA & 87.40 & 35.88\\
%  %& WD & 86.51 & 61.26\\
%  & $\mathbf{L_q}$ & \textbf{89.05} & 48.24 \\
%  & Vol & 87.96 & \textbf{81.97} \\
% \hline
\multirow{4}{*}{ResNet18 @ CIFAR10 - 0.8}
 %& NA & 39.67 & 16.67\\
 %& WD & 37.95 & 16.47\\
 & $\mathbf{L_q}$ & {47.50} & 22.58 \\
 & Vol(0.25, 0.5) & 45.56 & {38.16} \\
  &  - Binary & 13.01 & 10.00 \\
  &  - Ternary & \textbf{48.72} & \textbf{45.85} \\
\hline
\multirow{4}{*}{LSTMATT @ IMDB - 0.4}
 %& NA & 57.18 & 55.30\\
 %& WD & 60.99 & 55.38\\
 & $\mathbf{L_q}$ & 59.17 & 56.35 \\
 & Vol(0.25, 0.5) & \textbf{65.02} & 55.53 \\
 &  - Binary & 63.56 & \textbf{62.31} \\
 &  - Ternary & 63.05 & 60.20 \\
 \hline
\end{tabular}
}
\end{table}

%\begin{figure}[t!]   

\begin{figure*}[t!]
\begin{subfigure}{0.5\textwidth}
           \includegraphics[trim=0 0 0 0, clip, width=1\linewidth]{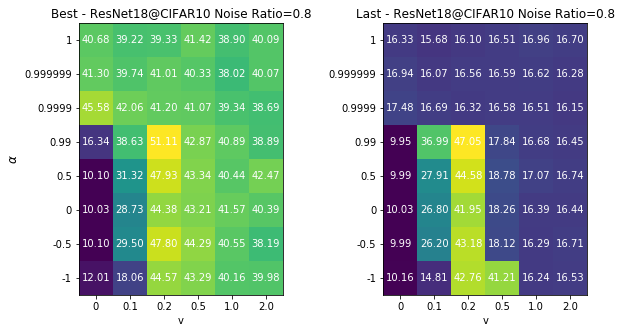}
    \vspace{-5.5mm}
    \caption{Grid search on CIFAR10 with ResNet18. We see that both the peak accuracy and the last accurary are more sensitive to volume than weight decay. Here, the best performance is achieved around $v=0.25,\ \alpha= 0.99$; we note a closer gridsearch in this region will result in even better performance. }
    \vspace{1mm}
    \label{fig:grid-search-04-main}
\end{subfigure}
\hfill
\begin{subfigure}{0.45\textwidth}
    \centering
    \includegraphics[width=0.49\linewidth]{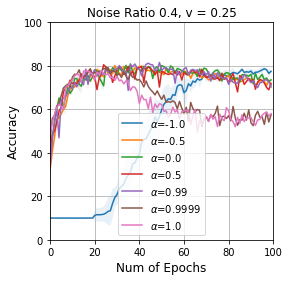}
    \includegraphics[width=0.49\linewidth]{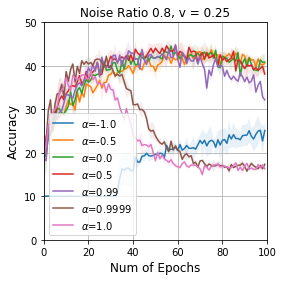}
    \vspace{-4mm}
    \caption{Training trajectory of ResNet18@CIFAR10 with label noise. We see that smaller $v$ leads to better robustness to the negative effect of mislabeling. We see that the baseline drop very quickly in performance after 20-th epoch due to learning the mislabeled points. On the other hand, a good combination of $v$ and $\alpha$ results in a training trajectory that drops very slowly or does not drop at all.}
    \label{fig:cifar10-label-noise}
\end{subfigure}
\vspace{-3mm}
\caption{Grid Search and training trajectory on CIFAR10 with ResNet18.}
\vspace{-3mm}
\end{figure*}

\begin{figure}[t!]   
    \centering
    \includegraphics[width=0.49\linewidth]{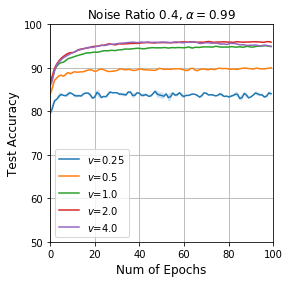}
    \includegraphics[width=0.49\linewidth]{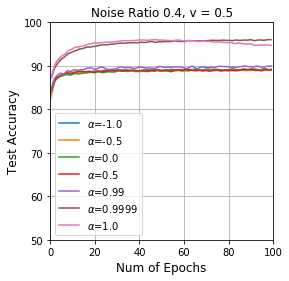}
    \includegraphics[width=0.49\linewidth]{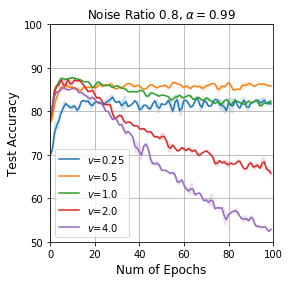}
    \includegraphics[width=0.49\linewidth]{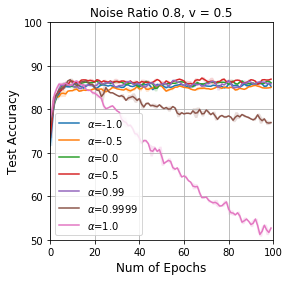}
    \vspace{-4mm}
    \caption{Test accuracy of DNN@MNIST with label noise}
    \label{fig:mnist-label-noise}
%\end{figure}
%\begin{figure}[t]
%\end{figure}
%\begin{figure}[t]
    \centering
    \includegraphics[width=0.49\linewidth]{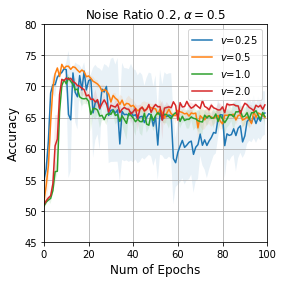}
    \includegraphics[width=0.49\linewidth]{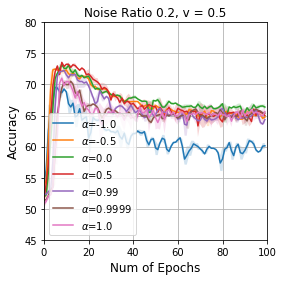}
    \includegraphics[width=0.49\linewidth]{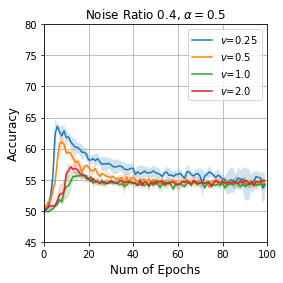}
    \includegraphics[width=0.49\linewidth]{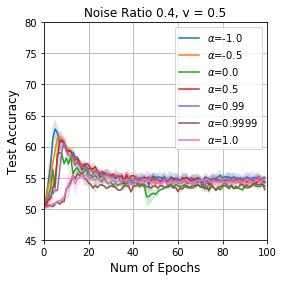}
    \vspace{-4mm}
    \caption{Test accuracy of LSTMATT@IMDB with label noise. We observe that tuning $v$ gives improvement in peak performance by upto $10\%$ absolute accuracy.}
    \label{fig:imdb-label-noise}
    \vspace{-1.5em}
\end{figure}

In this section, we show that the volumization can be more robust to label noises than weight decay. Label noise is an important topic in the recent machine learning research. We also compare against the previous state-of-the-art method (generalized cross entropy loss $\mathbf{L_q}$~\cite{zhang2018generalized}) with similar computational complexity on extreme noise rates. 
We use 2-layer DNN on MNIST~\cite{lecun1998gradient}, ResNet-18~\cite{he2016deep} on CIFAR-10~\cite{krizhevsky2009learning} and LSTMATT on IMDB dataset~\cite{maas-EtAl:2011:ACL-HLT2011}.
%For DNN, The dimension of hidden layer is 256.
For the LSTMATT~\cite{bahdanau2014neural}, we also use $300$-dimensional pretrained GloVe embedding~\cite{pennington2014glove}. %and set the dimension of hidden states to be 256.
We run all the cases for 100 epochs and report the best and last 10 epoch average of test accuracy trajectory.
We do not perform early stopping in order to showcase the strength of each method to prevent overfitting when no other measure against noisy label is taken.
$v$ is chosen from $\{0.25,0.5,1,2\}$.
We note that when $v$ is large, i.e. $v=2, 4$, the regularization of volumization is weak.
Experiments on Wider range of $v$ is included if needed.
From both CIFAR10 and IMDB results, it is observed that volumization significantly improves Best and Last score when label noise is large (CIFAR10-0.8, IMDB-0.4). While for the cases with smaller label noise, volumization's performance is compatible to weight decay. 

Our method is more similar to the loss function approach that limits the expressivity of the model. Previously proposed methods include $\mathbf{L_q}$~\cite{zhang2018generalized}), gambler's loss \cite{ziyin2019deep, ziyin2020learning} etc.. In table~\ref{tab: benchmark}, we compare volumization with the $\mathbf{L_q}$ method in preventing label noise. We focus on the case when the noise level is extreme, since this is where volumization is observed to be the most effective. We see that volumization achieves the state-of-the-art performance this level. Also shown are the performance of quantization using volumization. See section~\ref{exp: quantization} for more detail.
%In Table~\ref{table:label-noise}, we quantitatively compare the effect of volumization with weight decay under varying noise condition in MNIST, CIFAR10 and IMDB dataset.
%For each case, we present the best test accuracy in the trajectory as well as last 10 epochs average.

\vspace{-2mm}
\subsubsection{How to tune $V$ and $\alpha$}
\vspace{-1.5mm}

It is also beneficial to study how using volumization changes the training trajectories.
Figure~\ref{fig:mnist-label-noise}, \ref{fig:cifar10-label-noise} and \ref{fig:imdb-label-noise} shows the training trajectories on three datasets. To demonstrate the connection of our volumization with weight decay, weight cliping and elastic wall. We choose alpha from $\{-1,-0.5,0.,0.5, 0.99,0.9999,1 \}$.
We note again that when $\alpha$ is close to 1, the regularization is not significant and close to weight decay. when $\alpha = 0$, it is weight clipping and when $\alpha=-1$, it is elastic wall.
Take DNN@MNIST in Figure~\ref{fig:mnist-label-noise} as an example. In the first column of Figure~\ref{fig:mnist-label-noise}, we compare the effect of $v$ given $\alpha=0.99$.
Smaller $v$, i.e. $v < 1$ indicates stronger regularization and the overfitting is successfully prevented.
Larger $v$ has weaker regularization effect.
When label noise is large, i.e. 0.8, we see that only smaller $v$s prevents overfitting.
When label noise is small, i.e. 0.4, however, the smaller $v$s cause the underfitting.
In the second column in Figure~\ref{fig:mnist-label-noise}, we compare the effect of $\alpha$ given $v=0.5$.
Volumization prevent overfitting for almost all $\alpha$ under varying noise condition except $\alpha=1$, the no regularization case.
When label noise is large, clear regularization effect is observed when $\alpha < 1$ even for $\alpha$ close to 1 (the case where $\alpha=0.99$).
For the $\alpha$ getting exponentially closer to 1 ($\alpha=0.9999$), the regularization effect gets weaker and approximates to the no regularization case.

Figure~\ref{fig:cifar10-label-noise} presents the test accuracy trajectory of ResNet18 on CIFAR10.
Proper $v$ and $\alpha$ are important to make the model resistant to label noise.
We note that for the stronger label noise conditions, i.e. noise ratio is 0.8, stronger regularization should be applied.
Compared to the case where noise ratio is 0.4 and the primary hyperparameter $v$ is chosen to be $0.25$, for noise ratio 0.8 case we should choose $(v,\ \alpha)=(0.2, 0.5)$.
Figure~\ref{fig:imdb-label-noise} demonstrated the test accuracy trajectory of LSTMATT on IMDB dataset.
IMDB is a binary classification task so the label noise ratio should be smaller than 0.5 for this dataset.
We choose noise ratio 0.2 for moderate label noise and 0.4 for large label noise.
In this dataset, we observed that volumization could significantly improve the best performance of the trajectory.
Similar to previous observation, stronger label noise requires stronger regularization.
%We could see that in noise ratio 0.2 case, $v=0.5, \alpha=-1$ makes the model underfit the data, while in noise ratio 0.4 case, $v=0.5, \alpha=-1$ performance well compared to the no regularization baseline $\alpha=1$.
% However, the test accuracy trajectories of IMDB don't keep in a very high level.
% The reason may be credit to the different type of RNN network and dataset.

Based on the observations above, we conclude that $(v,\ \alpha)\in [0, +\infty)\times [-1, 1)$ (in the region of interest in Figure~\ref{fig:phase-diagram}) controls the strength of the regularization and can improve generalization.
Smaller $v$ and $\alpha$ indicates stronger regularization effect and vise versa.
Moreover, $v$ appears to be more sensitive and is considered the dominate hyperparameter for volumization.
$\alpha$, at the same time, is not very sensitive in most part of $[-1, 1)$, i.e. $[-1, -0.99]$, and could be considered as a secondary hyperparameter for fine-grained turning.

\begin{figure}[t]
    \centering
    \includegraphics[width=0.49\linewidth]{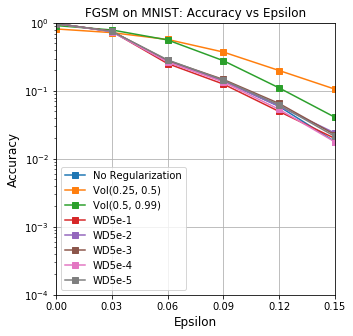}
    \includegraphics[width=0.49\linewidth]{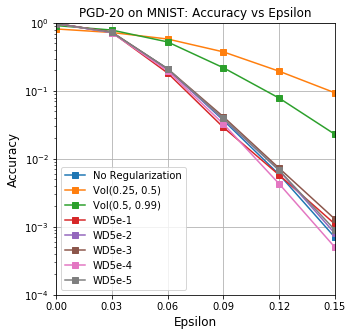}
    \vspace{-5.5mm}
    \caption{FGSM and PGD-20 attack on 2-layer-DNN@MNIST}
    \label{fig:attack-mnist}
    
%\end{figure}

%\begin{figure}[t]
\vspace{-0mm}
    \centering
    \includegraphics[width=0.49\linewidth]{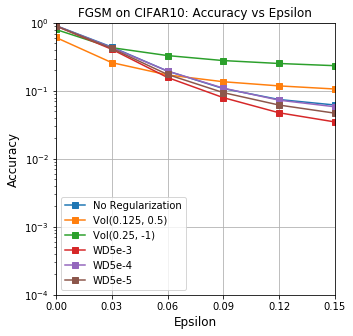}
    \includegraphics[width=0.49\linewidth]{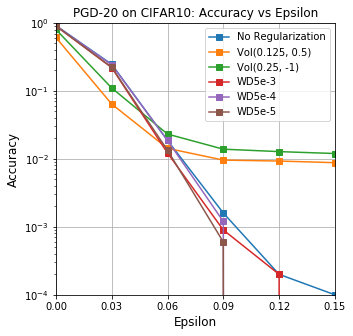}
    \vspace{-5.5mm}
    \caption{FGSM and PGD-20 attack on ResNet18@CIFAR10}
    
    \label{fig:attack-CIFAR10}
    \vspace{-1.5em}
\end{figure}
\vspace{-3mm}
\subsection{Robustness to Adversarial Attack}
\vspace{-1.5mm}
We also find that training with volumization makes models more robust to strong adversarial attack than weight decay.
This is related to Lipfhitz countinuity by the bounded eigen value of weight matrics
We perform simple Fast Gradient Sign Method (FGSM) attack~\cite{goodfellow2014explaining} and Projection Gradient Descent with 20 iterations (PGD-20) attack~\cite{madry2017towards} to 2-layer-DNN @ MNIST and ResNet18 @ CIFAR10.

Figure~\ref{fig:attack-mnist} and Figure~\ref{fig:attack-CIFAR10} compares weight decay and volumization under different attack strength in MNIST and CIFAR10, respectively.
For MNIST case, volumization are more robust for adversarial attack than weight decay. Actually, weight decay shows similar results to learning without any regularization.
Moreover, the case \emph{Vol(0.25, 0.5)} has surprisingly consistent performance under both naive FGSM and stronger PGD-20, while all weight decay case dropped significantly with stronger attack.

For ResNet18 @ CIFAR10 case, we still observe that volumization makes the models more robust to adversarial attacks, especially for adversarial examples with larger strength.
The weight decay cases perform even worse than that without regularization for ResNet18 @ CIFAR10 and decrease steadly under PGD-20 attack.
Interestingly, volumization are demonstrated to be ``saturated'' with the attack strength epsilon larger than 0.09.

\vspace{-3mm}
\subsection{Controlling Parameter Distribution and Weight Quantization}\label{exp: quantization}
\vspace{-1.5mm}

\begin{figure}[t]
%\centering

\begin{subfigure}[b]{0.24\linewidth}
    \centering
    \includegraphics[trim=0 0 0 0, clip,width=1\textwidth]{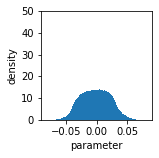}
    \vspace{-6mm}
    \caption{$V=1.2$}
    \label{fig: v004}
\end{subfigure}
\hfill
%\centering
\begin{subfigure}[b]{0.24\linewidth}
    \centering
    \includegraphics[trim=0 0 0 0, clip,width=1\textwidth]{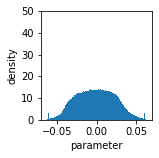}
    \vspace{-6mm}
    %\caption{Ternarize}
    \caption{$V=0.9$}
    \label{fig: v003}
\end{subfigure}
\hfill
\begin{subfigure}[b]{0.24\linewidth}
    \centering
    \includegraphics[trim=0 0 0 0, clip,width=1\textwidth]{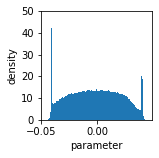}
    \vspace{-6mm}
    %\caption{Ternarize}
    \caption{$V=0.6$}
    \label{fig: v002}
\end{subfigure}
\hfill
\begin{subfigure}[b]{0.24\linewidth}
    \centering
    \includegraphics[trim=0 0 0 0, clip,width=1\textwidth]{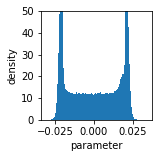}
    \vspace{-6mm}
    %\caption{Ternarize}
    \caption{$V=0.3$}
    \label{fig: v001}
\end{subfigure}

\begin{subfigure}[b]{0.24\linewidth}
    \centering
    \includegraphics[trim=0 0 0 0, clip,width=1\textwidth]{plots/quantization/v004a099.png}
    \vspace{-6mm}
    \caption{$V=1.2$}
    \label{fig: v0042}
\end{subfigure}
\hfill
%\centering
\begin{subfigure}[b]{0.24\linewidth}
    \centering
    \includegraphics[trim=0 0 0 0, clip,width=1\textwidth]{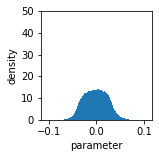}
    \vspace{-6mm}
    %\caption{Ternarize}
    \caption{$V=0.9$}
    \label{fig: v0032}
\end{subfigure}
\hfill
\begin{subfigure}[b]{0.24\linewidth}
    \centering
    \includegraphics[trim=0 0 0 0, clip,width=1\textwidth]{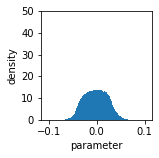}
    \vspace{-6mm}
    %\caption{Ternarize}
    \caption{$V=0.6$}
    \label{fig: v0022}
\end{subfigure}
\hfill
\begin{subfigure}[b]{0.24\linewidth}
    \centering
    \includegraphics[trim=0 0 0 0, clip,width=1\textwidth]{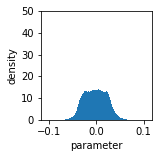}
    \vspace{-6mm}
    %\caption{Ternarize}
    \caption{$V=0.3$}
    \label{fig: v0012}
\end{subfigure}
\vspace{-3mm}
\caption{Controlling distribution of weight parameters of a two-layer network on MNIST. \textit{(a-d)}: $\alpha=0.99$; \textit{(e-h)}: $\alpha=0.9999$. All four networks performs similarly at $96\%$ accuracy. We note that, as $V$ decreases, peaks appear at $V$ and $-V$ and becomes higher, making these two points natural points of quantization. Such peak do not appear when either $\alpha$ or $V$ is large, meaning that both are very important at controlling the distribution of weight parameters. }\label{fig: distribution control}
%\vspace{-4mm}
%\end{figure}

%\begin{figure}[t]
%\centering

\begin{subfigure}[b]{0.25\textwidth}
    \centering
    \includegraphics[trim=0 0 0 0, clip,width=1\textwidth]{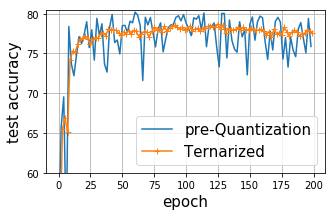}
    \vspace{-6mm}
    \caption{Ternarize}
    %\label{fig: depth, width256}
\end{subfigure}
\hfill
%\centering
\begin{subfigure}[b]{0.22\textwidth}
    \centering
    \includegraphics[trim=20 0 0 0, clip,width=1\textwidth]{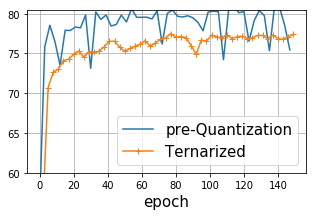}
    \vspace{-6mm}
    \caption{Binarize}
    %\label{fig: depth, width256}
\end{subfigure}
\vspace{-4mm}
\caption{Quantization of an LSTM network trained on the IMDB dataset. We see that the proposed method achieves quantization very smoothly, and the testing accuracy is also comparable to the original network.}\label{fig: quantize}
\vspace{-1.em}
\end{figure}

In this work, we show that the proposed method can also be used for controlling the distribution of weights of a network, and this can be directly used for quantizing neural networks, especially to binarize and to ternarize. See Figure~\ref{fig: distribution control}. We see that, volumization functions as an attractive force that pulls weight parameters towards $V$ and $-V$ at strength $\alpha$.

For demonstration, we perform a weight-only quantization \cite{guo2018survey} to $\pm V$. To binarize, we quantize weight $w\to V$ if $w>=0$ and $w\to -V$ otherwise. To ternarize, we use the following threshhold:
\begin{equation}
    w\to \begin{cases}
    V & \text{if } w > \frac{V}{2}\\
    0 & \text{if } \frac{-V}{2}\leq w \leq \frac{V}{2}\\
    V & \text{if } w < \frac{-V}{2}\\
    \end{cases}
\end{equation}
During training, we quantize every $2$ epochs, and continue training from this quantized network. While this quantization scheme is simple and ad hoc, its performance is quite good, resulting in a $32$-fold reduction in memory without incurring much drop in performance. One additional benefit resulting from quantizing with volumization is that it results naturally in a quantized network that is robust to the presence of label noise in the dataset. We plan to develop this quantization scheme in detail in a future work. See Figure~\ref{fig: quantize} for training trajectory of such quantization scheme, the quantized network achieves comparable performance to the original. 

Since volumization is robust to label noise, we expect that this method can be used to quantize a network that is simultaneously robust to label noise. We perform this experiment on the above mentioned setting on CIFAR-10 and IMDB when the noise is extreme ($0.8$ for CIFAR-10, $0.4$ for IMDB). See Table~\ref{tab: benchmark}. We set $V=0.1, \alpha=0.95$ in these settings with $lr=2e-4$. The proposed method does succeed in quantizing such networks. What is actually surprising is that such networks are benefited from the quantization scheme and the robustness is further improved. Achieving the state-of-the-art result on label noise problems at extreme noise levels. To our best knowledge, this is the first method for quantizing a neural network that defends against corrupted labels at the same time. We plan to study this in detail in a future work.

\vspace{-3mm}
\section{Conclusion}
\vspace{-1.5mm}
We propose \emph{volumization}, a novel regularization for deep neural networks as a generalization of weight decay.
\emph{Volumization} is proved in mild assumptions and shown to improve generalization better than weight decay in benchmark dataset.
We conducted extensive experiments including noisy label classification and adversarial attack to verify that volumization indeed has much wider applicability than weight decay, and performs significantly better when noise in learning is strong by preventing memorization of overparametrized neural networks. Moreover, We also showed that volumized can be used as a method to control the distribution of weight parameters, with quantization as a direct application, and it becomes the first method to be able to quantize neural networks and deal with label noise at the same time. There are many potential future works that the present work can lead to. We plan to extend the work on quantization to develop a binarization and ternarization method that is applicable for industrial tasks. The volumization technique (with the \textit{elastic wall hyperparameter}) can also be used to define a physical volume for neural networks, which may enable more study of neural networks from a theoretical physics point of view.

%\clearpage
%\bibliography{example_paper}

\begin{thebibliography}{40}
\providecommand{\natexlab}[1]{#1}
\providecommand{\url}[1]{\texttt{#1}}
\expandafter\ifx\csname urlstyle\endcsname\relax
  \providecommand{\doi}[1]{doi: #1}\else
  \providecommand{\doi}{doi: \begingroup \urlstyle{rm}\Url}\fi

\bibitem[Arjovsky et~al.(2017)Arjovsky, Chintala, and
  Bottou]{arjovsky2017wasserstein}
Arjovsky, M., Chintala, S., and Bottou, L.
\newblock Wasserstein gan.
\newblock \emph{arXiv preprint arXiv:1701.07875}, 2017.

\bibitem[Arora et~al.(2018)Arora, Cohen, and Hazan]{arora2018optimization}
Arora, S., Cohen, N., and Hazan, E.
\newblock On the optimization of deep networks: Implicit acceleration by
  overparameterization.
\newblock \emph{arXiv preprint arXiv:1802.06509}, 2018.

\bibitem[Bahdanau et~al.(2014)Bahdanau, Cho, and Bengio]{bahdanau2014neural}
Bahdanau, D., Cho, K., and Bengio, Y.
\newblock Neural machine translation by jointly learning to align and
  translate.
\newblock \emph{arXiv preprint arXiv:1409.0473}, 2014.

\bibitem[Belkin et~al.(2018)Belkin, Hsu, Ma, and Mandal]{belkin2018reconciling}
Belkin, M., Hsu, D., Ma, S., and Mandal, S.
\newblock Reconciling modern machine learning practice and the bias-variance
  trade-off.
\newblock \emph{arXiv preprint arXiv:1812.11118}, 2018.

\bibitem[Du et~al.(2018)Du, Zhai, Poczos, and Singh]{du2018gradient}
Du, S.~S., Zhai, X., Poczos, B., and Singh, A.
\newblock Gradient descent provably optimizes over-parameterized neural
  networks.
\newblock \emph{arXiv preprint arXiv:1810.02054}, 2018.

\bibitem[Gardner \& Derrida(1989)Gardner and Derrida]{gardner1989three}
Gardner, E. and Derrida, B.
\newblock Three unfinished works on the optimal storage capacity of networks.
\newblock \emph{Journal of Physics A: Mathematical and General}, 22\penalty0
  (12):\penalty0 1983, 1989.

\bibitem[Geiger et~al.(2019)Geiger, Spigler, d'Ascoli, Sagun, Baity-Jesi,
  Biroli, and Wyart]{geiger2019jamming}
Geiger, M., Spigler, S., d'Ascoli, S., Sagun, L., Baity-Jesi, M., Biroli, G.,
  and Wyart, M.
\newblock Jamming transition as a paradigm to understand the loss landscape of
  deep neural networks.
\newblock \emph{Physical Review E}, 100\penalty0 (1):\penalty0 012115, 2019.

\bibitem[Goodfellow et~al.(2016)Goodfellow, Bengio, and
  Courville]{goodfellow2016deep}
Goodfellow, I., Bengio, Y., and Courville, A.
\newblock \emph{Deep learning}.
\newblock 2016.

\bibitem[Goodfellow et~al.(2014)Goodfellow, Shlens, and
  Szegedy]{goodfellow2014explaining}
Goodfellow, I.~J., Shlens, J., and Szegedy, C.
\newblock Explaining and harnessing adversarial examples.
\newblock \emph{arXiv preprint arXiv:1412.6572}, 2014.

\bibitem[Gouk et~al.(2018)Gouk, Frank, Pfahringer, and
  Cree]{gouk2018regularisation}
Gouk, H., Frank, E., Pfahringer, B., and Cree, M.
\newblock Regularisation of neural networks by enforcing lipschitz continuity.
\newblock \emph{arXiv preprint arXiv:1804.04368}, 2018.

\bibitem[Guo(2018)]{guo2018survey}
Guo, Y.
\newblock A survey on methods and theories of quantized neural networks.
\newblock \emph{arXiv preprint arXiv:1808.04752}, 2018.

\bibitem[Hastie et~al.(2019)Hastie, Montanari, Rosset, and
  Tibshirani]{hastie2019surprises}
Hastie, T., Montanari, A., Rosset, S., and Tibshirani, R.~J.
\newblock Surprises in high-dimensional ridgeless least squares interpolation.
\newblock \emph{arXiv preprint arXiv:1903.08560}, 2019.

\bibitem[He et~al.(2015)He, Zhang, Ren, and Sun]{he2015delving}
He, K., Zhang, X., Ren, S., and Sun, J.
\newblock Delving deep into rectifiers: Surpassing human-level performance on
  imagenet classification.
\newblock In \emph{Proceedings of the IEEE international conference on computer
  vision}, pp.\  1026--1034, 2015.

\bibitem[He et~al.(2016)He, Zhang, Ren, and Sun]{he2016deep}
He, K., Zhang, X., Ren, S., and Sun, J.
\newblock Deep residual learning for image recognition.
\newblock In \emph{Proceedings of the IEEE conference on computer vision and
  pattern recognition}, pp.\  770--778, 2016.

\bibitem[Hendrycks et~al.(2019)Hendrycks, Mazeika, Kadavath, and
  Song]{hendrycks2019using}
Hendrycks, D., Mazeika, M., Kadavath, S., and Song, D.
\newblock Using self-supervised learning can improve model robustness and
  uncertainty.
\newblock In \emph{Advances in Neural Information Processing Systems}, pp.\
  15637--15648, 2019.

\bibitem[Kingma \& Ba(2014)Kingma and Ba]{journals/corr/KingmaB14_adam}
Kingma, D.~P. and Ba, J.
\newblock Adam: A method for stochastic optimization.
\newblock \emph{CoRR}, abs/1412.6980, 2014.
\newblock URL
  \url{http://dblp.uni-trier.de/db/journals/corr/corr1412.html#KingmaB14}.

\bibitem[Krizhevsky et~al.(2009)Krizhevsky, Hinton,
  et~al.]{krizhevsky2009learning}
Krizhevsky, A., Hinton, G., et~al.
\newblock Learning multiple layers of features from tiny images.
\newblock 2009.

\bibitem[Krogh \& Hertz(1992{\natexlab{a}})Krogh and
  Hertz]{krogh1992generalization}
Krogh, A. and Hertz, J.~A.
\newblock Generalization in a linear perceptron in the presence of noise.
\newblock \emph{Journal of Physics A: Mathematical and General}, 25\penalty0
  (5):\penalty0 1135, 1992{\natexlab{a}}.

\bibitem[Krogh \& Hertz(1992{\natexlab{b}})Krogh and Hertz]{krogh1992simple}
Krogh, A. and Hertz, J.~A.
\newblock A simple weight decay can improve generalization.
\newblock In \emph{Advances in neural information processing systems}, pp.\
  950--957, 1992{\natexlab{b}}.

\bibitem[Kuka{\v{c}}ka et~al.(2017)Kuka{\v{c}}ka, Golkov, and
  Cremers]{kukavcka2017regularization}
Kuka{\v{c}}ka, J., Golkov, V., and Cremers, D.
\newblock Regularization for deep learning: A taxonomy.
\newblock \emph{arXiv preprint arXiv:1710.10686}, 2017.

\bibitem[LeCun et~al.(1998)LeCun, Bottou, Bengio, and
  Haffner]{lecun1998gradient}
LeCun, Y., Bottou, L., Bengio, Y., and Haffner, P.
\newblock Gradient-based learning applied to document recognition.
\newblock \emph{Proceedings of the IEEE}, 86\penalty0 (11):\penalty0
  2278--2324, 1998.

\bibitem[Liang et~al.(2020)Liang, Liu, Ziyin, Salakhutdinov, and
  Morency]{liang2020think}
Liang, P.~P., Liu, T., Ziyin, L., Salakhutdinov, R., and Morency, L.-P.
\newblock Think locally, act globally: Federated learning with local and global
  representations.
\newblock \emph{arXiv preprint arXiv:2001.01523}, 2020.

\bibitem[Loshchilov \& Hutter(2017)Loshchilov and
  Hutter]{loshchilov2017decoupled}
Loshchilov, I. and Hutter, F.
\newblock Decoupled weight decay regularization.
\newblock \emph{arXiv preprint arXiv:1711.05101}, 2017.

\bibitem[Maas et~al.(2011)Maas, Daly, Pham, Huang, Ng, and
  Potts]{maas-EtAl:2011:ACL-HLT2011}
Maas, A.~L., Daly, R.~E., Pham, P.~T., Huang, D., Ng, A.~Y., and Potts, C.
\newblock Learning word vectors for sentiment analysis.
\newblock In \emph{Proceedings of the 49th Annual Meeting of the Association
  for Computational Linguistics: Human Language Technologies}, pp.\  142--150,
  Portland, Oregon, USA, June 2011. Association for Computational Linguistics.
\newblock URL \url{http://www.aclweb.org/anthology/P11-1015}.

\bibitem[Madry et~al.(2017)Madry, Makelov, Schmidt, Tsipras, and
  Vladu]{madry2017towards}
Madry, A., Makelov, A., Schmidt, L., Tsipras, D., and Vladu, A.
\newblock Towards deep learning models resistant to adversarial attacks.
\newblock \emph{arXiv preprint arXiv:1706.06083}, 2017.

\bibitem[Mei et~al.(2019)Mei, Misiakiewicz, and Montanari]{mei2019mean}
Mei, S., Misiakiewicz, T., and Montanari, A.
\newblock Mean-field theory of two-layers neural networks: dimension-free
  bounds and kernel limit.
\newblock \emph{arXiv preprint arXiv:1902.06015}, 2019.

\bibitem[M{\"u}ller et~al.(2019)M{\"u}ller, Kornblith, and
  Hinton]{muller2019does}
M{\"u}ller, R., Kornblith, S., and Hinton, G.~E.
\newblock When does label smoothing help?
\newblock In \emph{Advances in Neural Information Processing Systems}, pp.\
  4696--4705, 2019.

\bibitem[Nakkiran et~al.(2019)Nakkiran, Kaplun, Bansal, Yang, Barak, and
  Sutskever]{nakkiran2019deep}
Nakkiran, P., Kaplun, G., Bansal, Y., Yang, T., Barak, B., and Sutskever, I.
\newblock Deep double descent: Where bigger models and more data hurt.
\newblock \emph{arXiv preprint arXiv:1912.02292}, 2019.

\bibitem[Pennington et~al.(2014)Pennington, Socher, and
  Manning]{pennington2014glove}
Pennington, J., Socher, R., and Manning, C.~D.
\newblock Glove: Global vectors for word representation.
\newblock In \emph{Proceedings of the 2014 conference on empirical methods in
  natural language processing (EMNLP)}, pp.\  1532--1543, 2014.

\bibitem[Srivastava et~al.(2014)Srivastava, Hinton, Krizhevsky, Sutskever, and
  Salakhutdinov]{srivastava2014dropout}
Srivastava, N., Hinton, G., Krizhevsky, A., Sutskever, I., and Salakhutdinov,
  R.
\newblock Dropout: a simple way to prevent neural networks from overfitting.
\newblock \emph{The journal of machine learning research}, 15\penalty0
  (1):\penalty0 1929--1958, 2014.

\bibitem[Tan \& Le(2019)Tan and Le]{tan2019efficientnet}
Tan, M. and Le, Q.~V.
\newblock Efficientnet: Rethinking model scaling for convolutional neural
  networks.
\newblock \emph{arXiv preprint arXiv:1905.11946}, 2019.

\bibitem[Virmaux \& Scaman(2018)Virmaux and Scaman]{virmaux2018lipschitz}
Virmaux, A. and Scaman, K.
\newblock Lipschitz regularity of deep neural networks: analysis and efficient
  estimation.
\newblock In \emph{Advances in Neural Information Processing Systems}, pp.\
  3835--3844, 2018.

\bibitem[Wu et~al.(2017)Wu, Zhang, and Xu]{wu2017tiny}
Wu, J., Zhang, Q., and Xu, G.
\newblock Tiny imagenet challenge.
\newblock Technical report, Technical report, Stanford University, 2017. 2017.

\bibitem[Zhang et~al.(2017)Zhang, Bengio, Hardt, Recht, and
  Vinyals]{Zhang_rethink}
Zhang, C., Bengio, S., Hardt, M., Recht, B., and Vinyals, O.
\newblock Understanding deep learning requires rethinking generalization.
\newblock 2017.
\newblock URL \url{https://arxiv.org/abs/1611.03530}.

\bibitem[Zhang et~al.(2018{\natexlab{a}})Zhang, Liao, Rakhlin, Miranda,
  Golowich, and Poggio]{zhang2018theory}
Zhang, C., Liao, Q., Rakhlin, A., Miranda, B., Golowich, N., and Poggio, T.
\newblock Theory of deep learning iib: Optimization properties of sgd.
\newblock \emph{arXiv preprint arXiv:1801.02254}, 2018{\natexlab{a}}.

\bibitem[Zhang et~al.(2018{\natexlab{b}})Zhang, Wang, Xu, and
  Grosse]{zhang2018three}
Zhang, G., Wang, C., Xu, B., and Grosse, R.
\newblock Three mechanisms of weight decay regularization.
\newblock \emph{arXiv preprint arXiv:1810.12281}, 2018{\natexlab{b}}.

\bibitem[Zhang \& Sabuncu(2018)Zhang and Sabuncu]{zhang2018generalized}
Zhang, Z. and Sabuncu, M.
\newblock Generalized cross entropy loss for training deep neural networks with
  noisy labels.
\newblock In \emph{Advances in neural information processing systems}, pp.\
  8778--8788, 2018.

\bibitem[Ziyin et~al.()Ziyin, Wang, Liang, Salakhutdinov, Morency, and
  Ueda]{ziyin2019deep}
Ziyin, L., Wang, Z., Liang, P.~P., Salakhutdinov, R., Morency, L.-P., and Ueda,
  M.
\newblock Deep gamblers: Learning to abstain with portfolio theory.

\bibitem[Ziyin et~al.(2020{\natexlab{a}})Ziyin, Chen, Wang, Liang,
  Salakhutdinov, Morency, and Ueda]{ziyin2020learning}
Ziyin, L., Chen, B., Wang, R., Liang, P.~P., Salakhutdinov, R., Morency, L.-P.,
  and Ueda, M.
\newblock Learning not to learn in the presence of noisy labels.
\newblock \emph{arXiv preprint arXiv:2002.06541}, 2020{\natexlab{a}}.

\bibitem[Ziyin et~al.(2020{\natexlab{b}})Ziyin, Wang, and
  Ueda]{ziyin2020laprop}
Ziyin, L., Wang, Z.~T., and Ueda, M.
\newblock Laprop: a better way to combine momentum with adaptive gradient.
\newblock \emph{arXiv preprint arXiv:2002.04839}, 2020{\natexlab{b}}.

\end{thebibliography}

\bibliographystyle{icml2020}

\appendix
\onecolumn
\section*{Appendix}

%\subsection{Behavior of Volumization for Gaussian distribution}
%The setting is the same as the theory section.  See Figure~\ref{fig:theo gaussian}. We see that searching for $V$ gives improvement over without volumization. 
%\begin{figure}
%    \centering
%    \includegraphics[width=200pt]{plots/gaussian.png}
%    \vspace{-8mm}
%    \caption{Behavior of volumization when the true parameter distribution obeys $Gaussian(0, 0.25)$. While this is harder to solve analytically, We observe similar trend to the case when the true distribution is uniform.}
%    \label{fig:theo gaussian}
%\end{figure}

%Since the analytical solution to this is very ugly and not very insightful, we do not show it here. To linear order in $\sigma$, we have $\frac{\sigma}{a}y=0$, and the solution is $y=a$, which is what we expect since we know the optimal solution lies in the range $(-a,a)$, and so maybe the best strategy is to confine the solution to be such. At $V=a$, the generalization error is
%\begin{equation}
%    b^2(a) = \frac{3\sigma}{4a}\sigma^2
%\end{equation}
%and for $\sigma\leq a$, $ b^2(a) < \sigma^2 < b^2$. 

\section{Volumized SGD and ADAM}

In this section, we provide the pseudo-code for volumized SGD, Adam\cite{journals/corr/KingmaB14_adam}, and LaProp \cite{ziyin2020laprop}. See Algorithm \ref{alg:sgd}, \ref{alg:adam}, \ref{alg:laprop} respectively. We recommend using LaProp for its demonstrated better ability to initialize training and its better stability.

\begin{algorithm}
	\caption{volumized SGD }
	\label{alg:sgd}
	\begin{algorithmic}
		\STATE {\bfseries Input:} $x_1 \in \mathbb{R}^d$, learning rate $\{\lambda_t\}_{t=1}^T$, decay parameters $0 \leq \mu < 1$. Set $m_{0} = 0$, $n_{0} = 0  $.
		%\FOR{$t=1$ {\bfseries to} $T$}
		%\STATE Draw a sample $s_t$ from $\mathbb{P}$.
		
		\STATE $g_{t} = \nabla_\theta \ell(\theta_{t-1})$
		%\STATE $n_t = \nu n_{t-1} + (1-\nu) g_{t}^2$
		\STATE $m_t = \mu m_{t-1} + g_t  $
		%\STATE \#\#Following lines are the differences
		%\STATE $\mu_t = \beta_2 \mu_{t-1} + (1 - \beta_2) g_{t}$
		
		\STATE $w_{t+1} = w_t - \lambda_t  \times  m_t $
		\IF{$|w_{t+1}| > V$}
		\STATE \quad\quad $w_{t+1}= w_{i, t+1} + (1-\alpha)(V \times \texttt{sgn}({w}_{t+1}) -{w}_{t+1})$
		\STATE \quad\quad $m_{t} = \alpha \times m_{ t}$
		\ENDIF
	\end{algorithmic}
\end{algorithm}

\begin{algorithm}
	\caption{volumized Adam }
	\label{alg:adam}
	\begin{algorithmic}
		\STATE {\bfseries Input:} $x_1 \in \mathbb{R}^d$, learning rate $\{\lambda_t\}_{t=1}^T$, decay parameters $0 \leq \mu < 1,\ 0 \leq \nu < 1 ,\ \epsilon \ll 1$. Set $m_{0} = 0$, $n_{0} = 0  $.
		%\FOR{$t=1$ {\bfseries to} $T$}
		%\STATE Draw a sample $s_t$ from $\mathbb{P}$.
		
		\STATE $g_{t} = \nabla_\theta \ell(\theta_{t-1})$
		\STATE $n_t = \nu n_{t-1} + (1-\nu) g_{t}^2$
		\STATE $m_t = \mu m_{t-1} + (1-\mu) g_t  $
		%\STATE \#\#Following lines are the differences
		%\STATE $\mu_t = \beta_2 \mu_{t-1} + (1 - \beta_2) g_{t}$
		
		\STATE $w_{t+1} = w_t - \lambda_t  \times  \frac{ g_{t}}{\sqrt{n_t/c_n} + \epsilon} m_t / c_m $
		\IF{$|w_{t+1}| > V$}
		\STATE \quad\quad $w_{t+1}= w_{i, t+1} + (1-\alpha)(V \times \texttt{sgn}({w}_{t+1}) -{w}_{t+1})$
		\STATE \quad\quad $m_{t} = \alpha \times m_{ t}$
		\ENDIF
	\end{algorithmic}
\end{algorithm}

\begin{algorithm}
	\caption{volumized LaProp }
	\label{alg:laprop}
	\begin{algorithmic}
		\STATE {\bfseries Input:} $x_1 \in \mathbb{R}^d$, learning rate $\{\lambda_t\}_{t=1}^T$, decay parameters $0 \leq \mu < 1,\ 0 \leq \nu < 1 ,\ \epsilon \ll 1$. Set $m_{0} = 0$, $n_{0} = 0  $.
		%\FOR{$t=1$ {\bfseries to} $T$}
		%\STATE Draw a sample $s_t$ from $\mathbb{P}$.
		
		\STATE $g_{t} = \nabla_\theta \ell(\theta_{t-1})$
		\STATE $n_t = \nu n_{t-1} + (1-\nu) g_{t}^2$
		\STATE $m_t = \mu m_{t-1} + (1-\mu) \frac{ g_{t}}{\sqrt{n_t/c_n} + \epsilon} $
		%\STATE \#\#Following lines are the differences
		%\STATE $\mu_t = \beta_2 \mu_{t-1} + (1 - \beta_2) g_{t}$
		
		\STATE $w_{t+1} = w_t - \lambda_t m_t / c_m $
		\IF{$|w_{t+1}| > V$}
		\STATE \quad\quad $w_{t+1}= w_{i, t+1} + (1-\alpha)(V \times \texttt{sgn}({w}_{t+1}) -{w}_{t+1})$
		\STATE \quad\quad $m_{t} = \alpha \times m_{ t}$
		\ENDIF
	\end{algorithmic}
\end{algorithm}

\section{Grid Search of $(v, \alpha)$ and Default Parameters}

\begin{figure}
    \centering
    \includegraphics[width=0.8\textwidth]{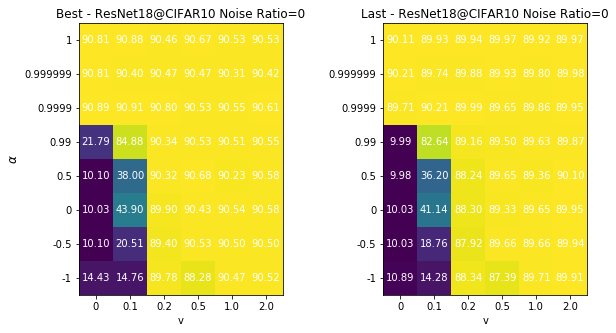}
    \vspace{-5.5mm}
    \caption{Grid search Best and Last scores on ResNet18@CIFAR10 with no label noise}
    \vspace{1mm}
    \label{fig:grid-search-0}
%\end{figure}

%\begin{figure}
    \centering
    \includegraphics[width=0.8\textwidth]{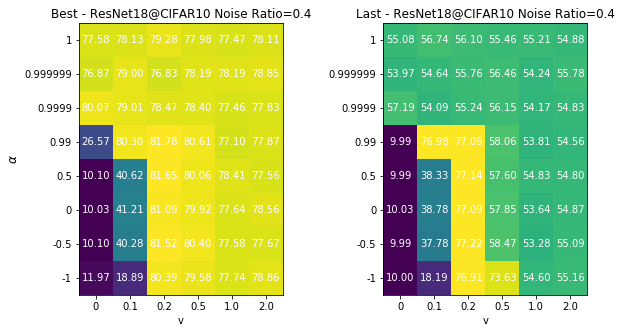}
    \vspace{-5.5mm}
    \caption{Grid search Best and Last scores on ResNet18@CIFAR10 with noise ratio 0.4}
    \vspace{1mm}
    \label{fig:grid-search-04}
%\end{figure}

%\begin{figure}
    \centering
    \includegraphics[width=0.8\textwidth]{plots/grid-search-cifar10-nr0_8.png}
    \vspace{-5.5mm}
    \caption{Grid search Best and Last scores on ResNet18@CIFAR10 with noise ratio 0.8}
    \vspace{1mm}
    \label{fig:grid-search-08}
\end{figure}

\begin{figure}[t]
\begin{subfigure}{0.475\textwidth}
        %\centering
    \includegraphics[width=\linewidth]{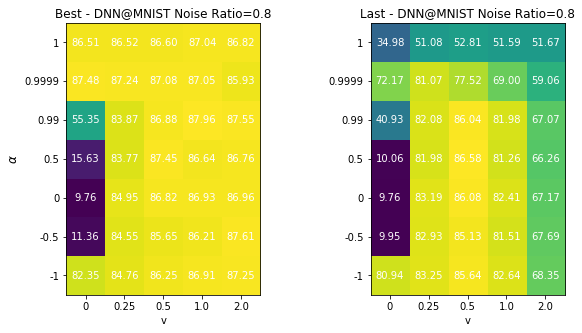}
    \vspace{-5.5mm}
    \caption{Grid search on MNIST with noise ratio 0.8}
    \vspace{1mm}
    \label{fig:mnist-grid}
\end{subfigure}
\hfill
\begin{subfigure}{0.475\textwidth}
           % \centering
    \includegraphics[width=\linewidth]{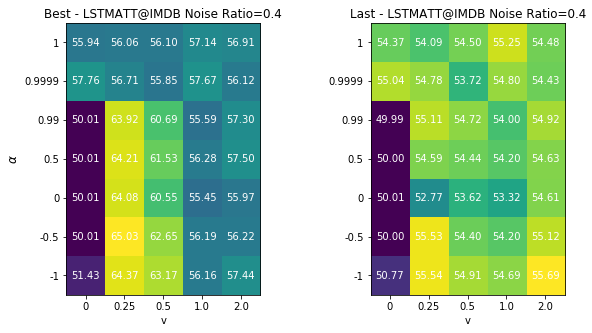}
    \vspace{-5.5mm}
    \caption{Grid search on IMDB with noise ratio 0.4}
    \vspace{1mm}
    \label{fig:imdb-grid}
\end{subfigure}
\vspace{-3mm}
\caption{Grid search Best and Last scores on IMDB and IMDB. We see that voluzamition has different effect on the two tasks. For IMDB, the peak performance is improved by volumization, and for MNIST, the final performance is improved.}
\end{figure}

We grid search the hyperparameters $(v, \alpha)$ on ResNet18 @ CIFAR10 under different label noise levels. Each $(v, \alpha)$ result is the average score of three runs. Figure~\ref{fig:grid-search-0}, Figure~\ref{fig:grid-search-04} and Figure~\ref{fig:grid-search-08} present the Best and Last score in the $(v, \alpha)$ phase diagram under no label noise, noise ratio 0.4 and 0.8. For $v=0$ case where volumization is actually weight decay, we see from smaller $\alpha \leq 0.99$, i.e. larger weight decay strength limits the learning ability of the neural network. However, in over-regularized $\alpha \leq 0.99$ cases, increasing the volume $v$ will improve the performance significantly, and then converge to the no regularization case. In all label noise condition, the most suitable hyperparameter $(v, \alpha)$ locates in where $(0, \alpha)$ over-regularizes the neural network and $(v, \alpha)$ finds the proper regularization. This suggests that the generalization performance is approximately convex in $v$, and our grid search indicates that $v$ is easier to search for than the weight decay $\alpha$, and $v$ is much more effective at constraining a neural network.

In Figure \ref{fig:mnist-grid} and \ref{fig:imdb-grid}, we show the grid search on MNIST and IMDB. We notice that the effect of volumization seems different on different datasets or architectures. For IMDB, the peak performance is improved by volumization, and for MNIST, the final performance is improved.
Interestingly, for two-layer-DNN @ MNIST, the $v = 0, \alpha=-1$ case is also learnable. This might be the result of the even time step symmetry.

\section{Similar Property on Other Adaptive Optimizers}
We also grid search the parameters $(v,\alpha)$ in aforementioned three cases using another adaptive optimizer Laprop~\cite{ziyin2020laprop}.

\begin{figure}
\begin{subfigure}{0.3\textwidth}
        %\centering
    \includegraphics[width=\linewidth]{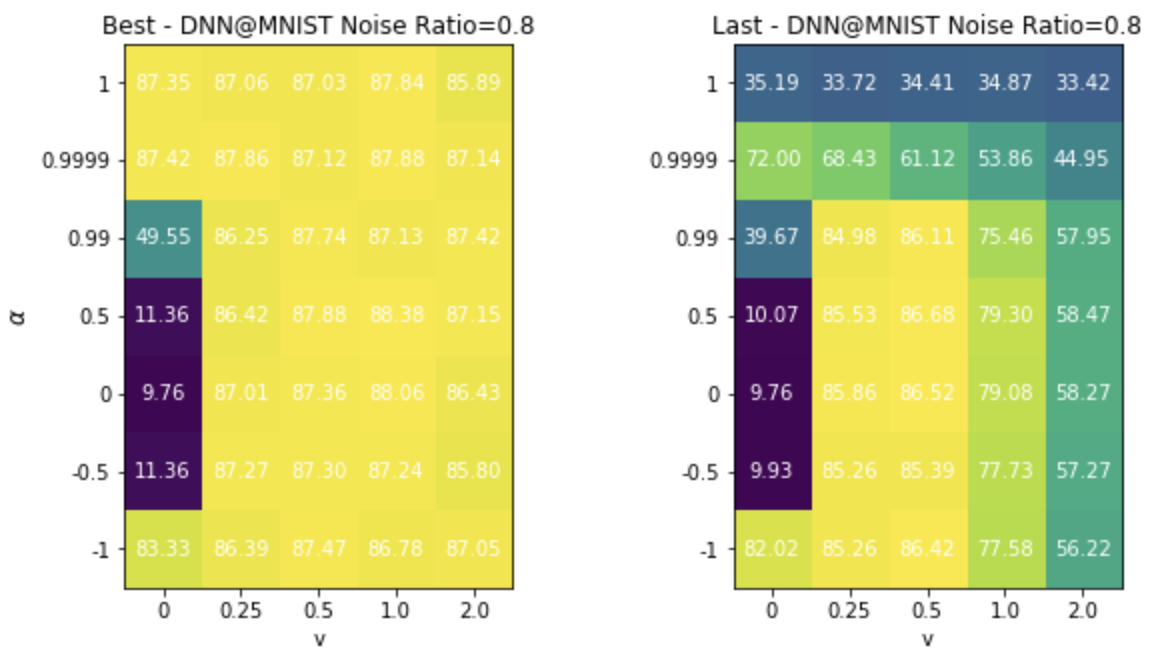}
    \vspace{-5.5mm}
    \caption{volumized Laprop in MNIST}
    \vspace{1mm}
\end{subfigure}
\hfill
\begin{subfigure}{0.3\textwidth}
        %\centering
    \includegraphics[width=\linewidth]{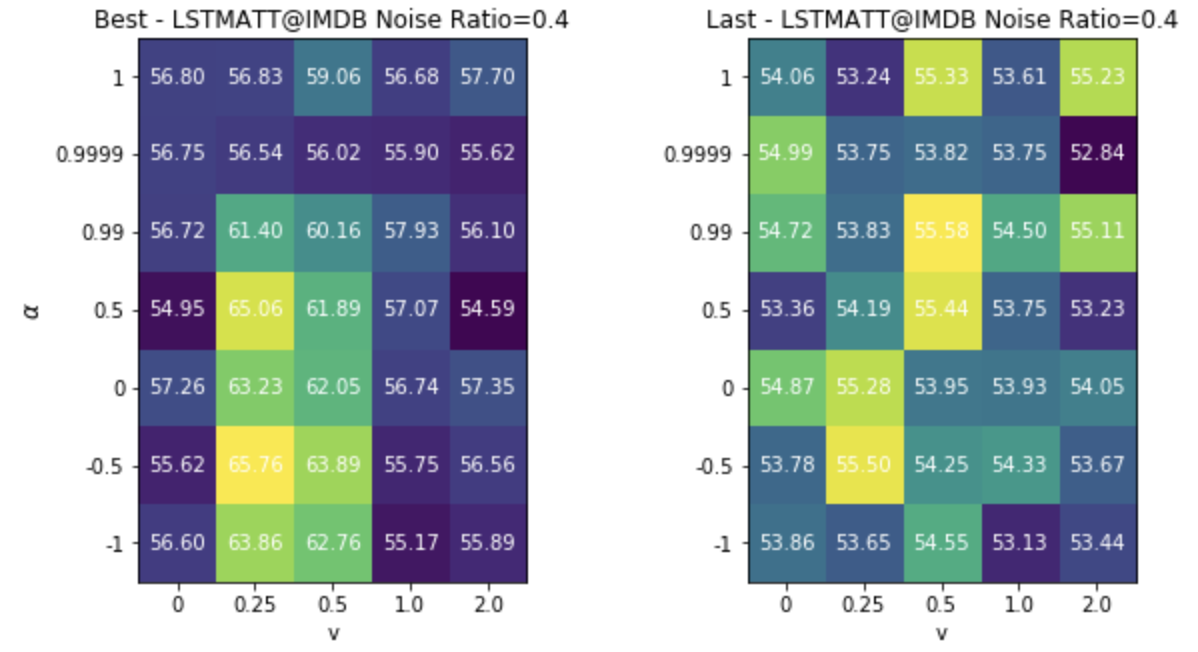}
    \vspace{-5.5mm}
    \caption{volumized Laprop in IMDB}
    \vspace{1mm}
\end{subfigure}
\hfill
\begin{subfigure}{0.3\textwidth}
        %\centering
    \includegraphics[width=\linewidth]{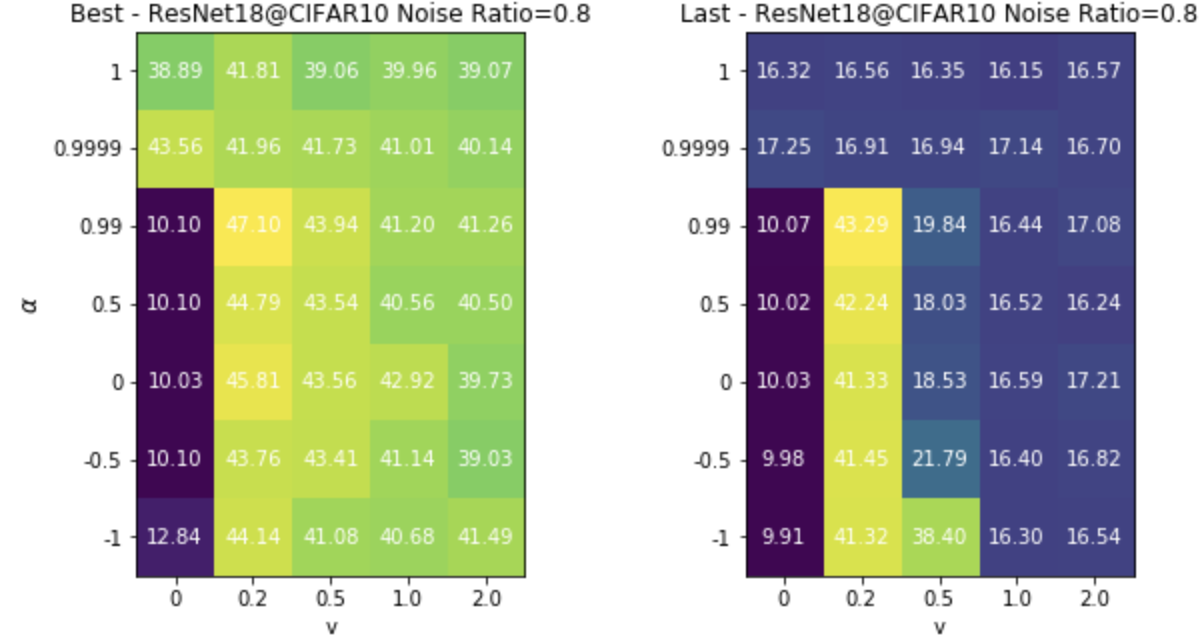}
    \vspace{-5.5mm}
    \caption{volumized Laprop in CIFAR10}
    \vspace{1mm}
\end{subfigure}
\caption{Grid search Best and Last scores optimized by volumized Laprop on MNIST, IMDB and CIFAR10 with the largest label noise level. For MNIST and CIFAR10, the Last scores are improved, for IMDB, the Best scores are improved}\label{fig:laprop-grid-search}
\end{figure}

Figure~\ref{fig:laprop-grid-search} demonstrated the grid search results. We could see the observation on Adam is invariant to the Laprop optimizer.

%%%%%%%%%%%%%%%%%%%%%%%%%%%%%%%%%%%%%%%%%%%%%%%%%%%%%%%%%%%%%%%%%%%%%%%%%%%%%%%
%%%%%%%%%%%%%%%%%%%%%%%%%%%%%%%%%%%%%%%%%%%%%%%%%%%%%%%%%%%%%%%%%%%%%%%%%%%%%%%
% DELETE THIS PART. DO NOT PLACE CONTENT AFTER THE REFERENCES!
%%%%%%%%%%%%%%%%%%%%%%%%%%%%%%%%%%%%%%%%%%%%%%%%%%%%%%%%%%%%%%%%%%%%%%%%%%%%%%%
%%%%%%%%%%%%%%%%%%%%%%%%%%%%%%%%%%%%%%%%%%%%%%%%%%%%%%%%%%%%%%%%%%%%%%%%%%%%%%%
\section{Proof of Theorems}
We first note that the original solution $u_i$ is shifted to $u_i' = u_i + \eta_i$, where $U_i \sim Unif(-a, a)$, and $\eta_i \sim Unif(-\sigma, \sigma)$. The variance for $\eta$ is $\sigma^2/3$.

By elementary probability theory, the support for $u_i'$ is on $(-a-\sigma, a+\sigma)$, and the probability density of $u_i'$ is given by
\begin{equation}
    f(u') = \begin{cases}
        \frac{1}{4a\sigma}u' + \frac{a + \sigma}{4a\sigma}  \quad\quad &\text{if } -a - \sigma \leq  u_i \leq  -a + \sigma \\
        
        \frac{1}{2a} \quad\quad &\text{if } -a + \sigma< u_i <  a - \sigma \\ 
        
        \frac{-1}{4a\sigma}u' + \frac{a + \sigma}{4a\sigma}  \quad\quad &\text{if } -a - \sigma \leq  u_i \leq  -a + \sigma \\
        
    \end{cases}
\end{equation}
for all $i$. When we use volumization of volume $V$, the parameter $w_i$ converges to $u_i'$ if $u_i' \in (-V, V)$, and $w_i$ converges to the boundary at $\pm V$ otherwise. If $u_i' \in (-V, V)$, then the error from a single parameter $w_i$ is given by $\frac{\sigma^2}{3N}$, and this happens with probability
\begin{equation}
    Pr(u' < V) = \begin{cases}

        \frac{V}{a} \quad\quad &\text{if } V  <  a - \sigma \\ 
        
        1 - \frac{1}{4a\sigma} (V-a -\sigma)^2 \quad\quad &\text{if } a - \sigma \leq  V \leq  a + \sigma \\
        
        1 \quad\quad &\text{if } V > a + \sigma\\
        
    \end{cases}
\end{equation}
When $u_i' \notin (-V,\ V)$, the contribution to the generalization error is then $\frac{1}{N}(|u_i| - V)^2$, when we take expectation over $u$, this is equal to $\frac{(a-V)^2}{12N}$, and this happens with $1$ minus the above probability.\footnote{from this point on, we drop the dummy factor $12N$ since the text becomes clearer without them.} %To prove the first theorem, we set $V= a-\sigma$, and the generalization error is this case is easy to find:
%\begin{equation}
%    12N b^2(V) = \frac{a-\sigma}{a}{\sigma^2} + \sigma a 
%\end{equation}

To summarize, the generalization error is\footnote{notice that we are treating $b^2$ as a function of $V$.}
\begin{equation}
     b^2(V) = \begin{cases}

        \frac{V}{a}\sigma^2 +  \frac{1}{a} (a-V)^3 \quad\quad &\text{if } V  <  a - \sigma \\ 
        
        c_V\sigma^2 + (1 - c_V)(a-V)^2 \quad\quad &\text{if } a - \sigma \leq  V \leq  a + \sigma \\
        
        \sigma^2 \quad\quad &\text{if } V > a + \sigma\\
        
    \end{cases}
\end{equation}
where $c_V = 1 - \frac{1}{4a\sigma} (V-a -\sigma)^2 $; we can notice that $b^2(a-\sigma) = b^2(a+\sigma) = b^2$, and for $V \in (a-\sigma, a+\sigma)$, $b^2(V)< b^2$. Using $V$ in this range, we expect a gain in generalization performance. Formally, we can prove the following inequality for $a-\sigma< V < a + \sigma$:
\begin{equation}
    b^2(V) = c_V\sigma^2 + (1 - c_V)(a-V)^2 < b^2 
\end{equation}
To show this, we note that $0 \leq|V-a|\leq \sigma$, and so 
\begin{align}
    b^2(V)& \leq c_V\sigma^2 + (1 - c_V)\sigma^2\\
    &= \sigma^2/3
\end{align}
with equality only if $V = a\pm \sigma$, which is not in the range under consideration, and we are done.

We now take derivative w.r.t $V$ and set to $0$ in the region $V\in (a-\sigma, a + \sigma)$ to obtain
\begin{equation}
    -\frac{\sigma}{2a}(y-\sigma) + \frac{1}{2a\sigma}(y-\sigma)y^2 +  \frac{1}{2a\sigma}(y-\sigma)^2y = 0
\end{equation}
where we defined $y=V-a$. This has two solutions at $V= a -\frac{b}{2}$, which is the global minimum, and $V=a+b$, which is a local maximum. Thus, the optimal solution $V = a -\frac{b}{2}$ yields generalization error
\begin{equation}
    b^2\bigg(a-\frac{\sigma}{2}\bigg) = \bigg( 1 -\frac{27\sigma}{64a} \bigg)\frac{\sigma^2}{3} < \frac{\sigma^2}{3}
\end{equation}
for $\sigma \leq a$.

Of course, we may also take the limit $\sigma^2 \to 0$ from above to obtain the generalization error in the absence of noise. The error is
\begin{equation}
     b^2(V) = \begin{cases}

        \frac{1}{a} (a-V)^3 \quad\quad &\text{if } V  \leq a \\ 
        
        0 &\text{if } V > a\\
        
    \end{cases}
\end{equation}
Interestingly enough, when $V$ is chosen to limit the complexity of the model beyong necessity, the effect to generalization error is only affect by a term of order $O(a^2)$, and when $a$ is far smaller than $1$, the negative effect is very small.

\clearpage
\vspace{-3mm}
\section{Weight Distribution of a Volumized ResNet18}
\begin{figure}[b!]
    \centering
    \includegraphics[width=0.8\textwidth]{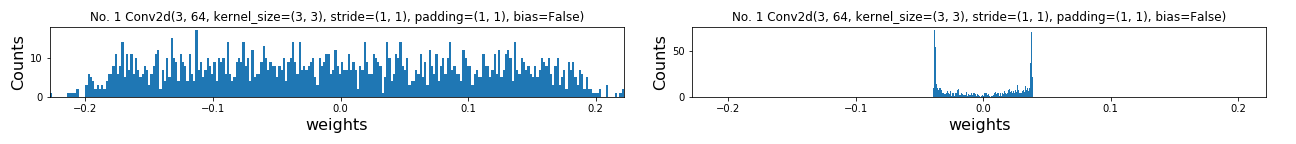}
    \includegraphics[width=0.8\textwidth]{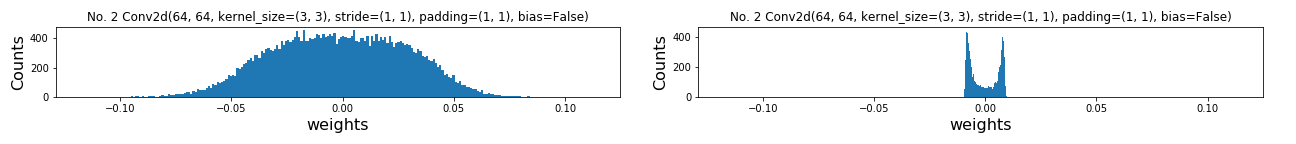}
    \includegraphics[width=0.8\textwidth]{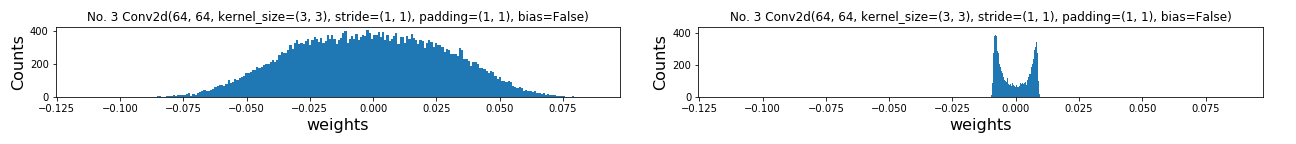}
    \includegraphics[width=0.8\textwidth]{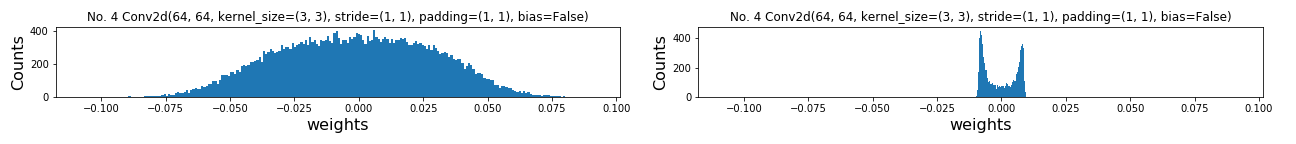}
    \includegraphics[width=0.8\textwidth]{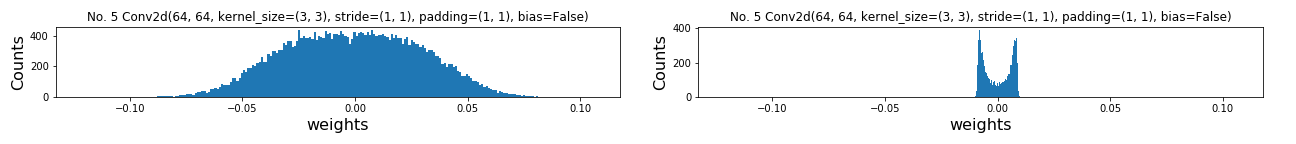}
    \includegraphics[width=0.8\textwidth]{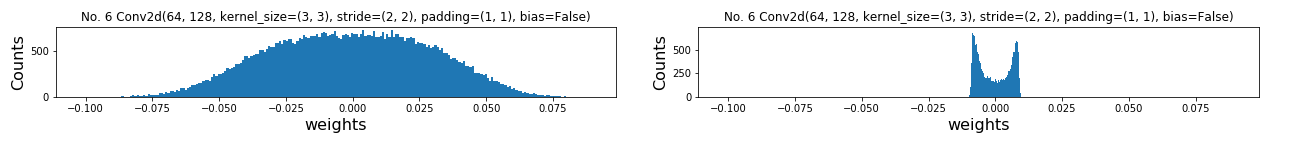}
    \includegraphics[width=0.8\textwidth]{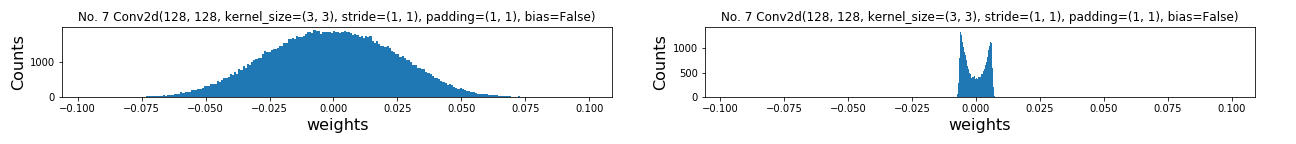}
    \includegraphics[width=0.8\textwidth]{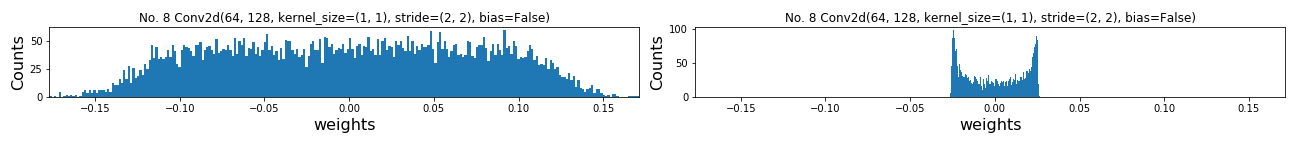}
    \includegraphics[width=0.8\textwidth]{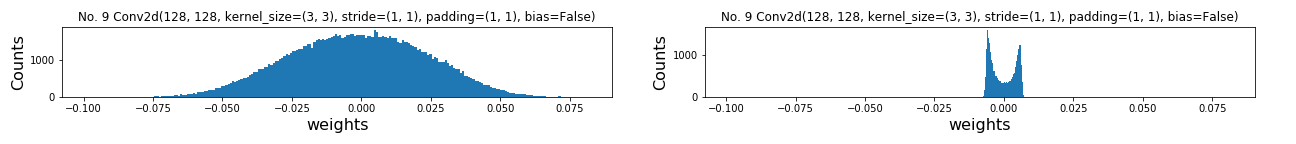}
    \vspace{-5.5mm}
    \caption{Weight distribution of ResNet18 trained on CIFAR10 with label noise ratio 0.4. Left: non-volumized; Right: volumized with $(v,\alpha)=(0.2, 0.99)$}
    \label{fig:weight-dist-0.4-1}
    \vspace{1mm}
\end{figure}

\begin{figure}
    \centering
    \includegraphics[width=0.8\textwidth]{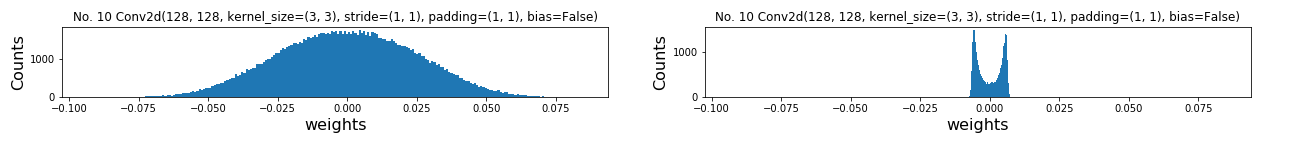}
    \includegraphics[width=0.8\textwidth]{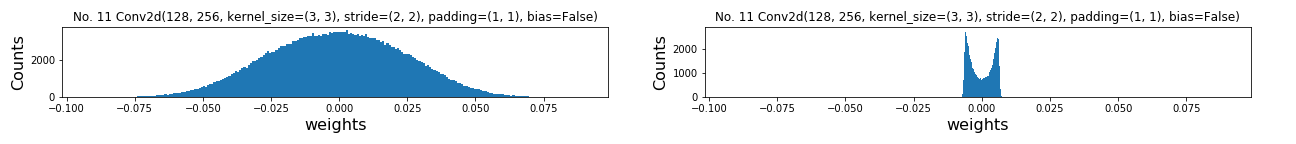}
    \includegraphics[width=0.8\textwidth]{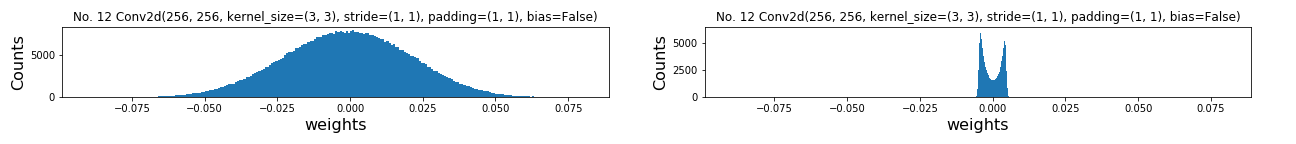}
    \includegraphics[width=0.8\textwidth]{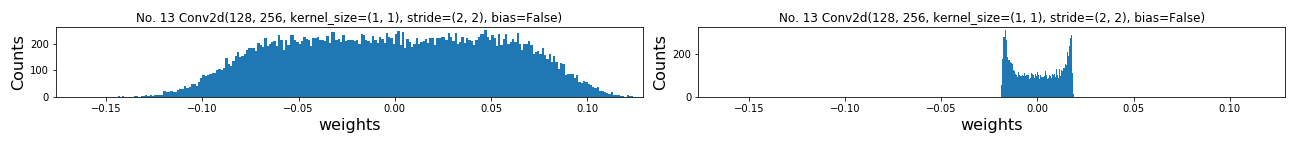}
    \includegraphics[width=0.8\textwidth]{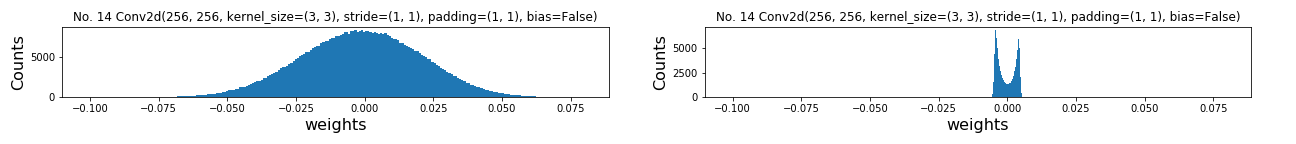}
    \includegraphics[width=0.8\textwidth]{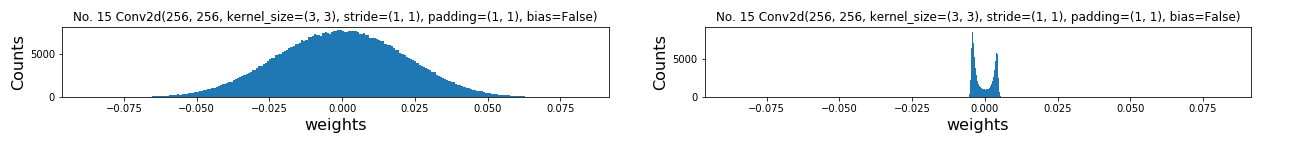}
    \includegraphics[width=0.8\textwidth]{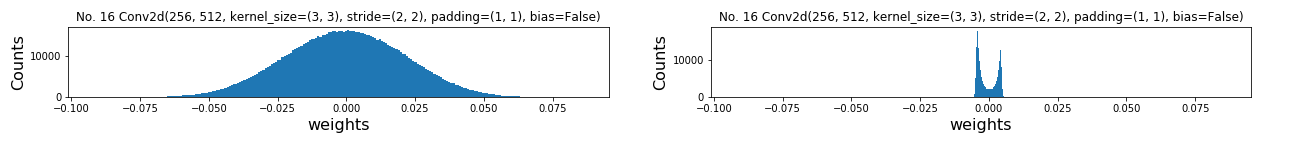}
    \includegraphics[width=0.8\textwidth]{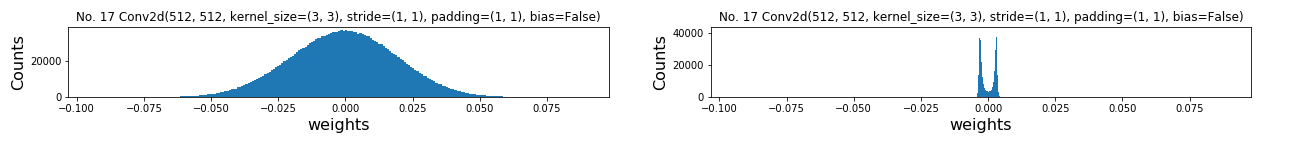}
    \includegraphics[width=0.8\textwidth]{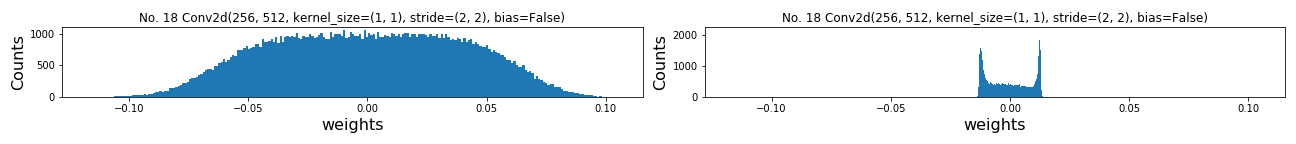}
    \includegraphics[width=0.8\textwidth]{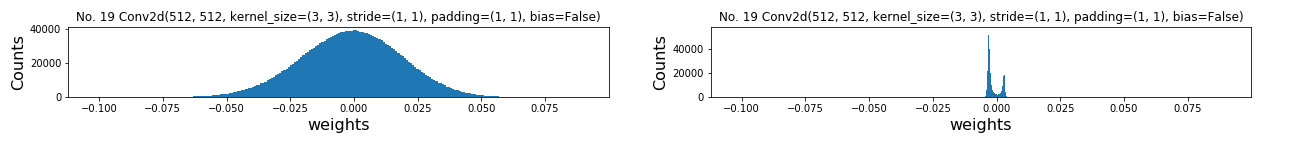}
    \includegraphics[width=0.8\textwidth]{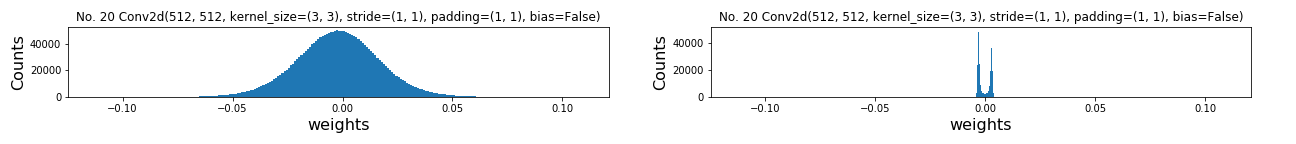}
    \includegraphics[width=0.8\textwidth]{plots/weight-dist/nr04-20conv.png}
    \includegraphics[width=0.8\textwidth]{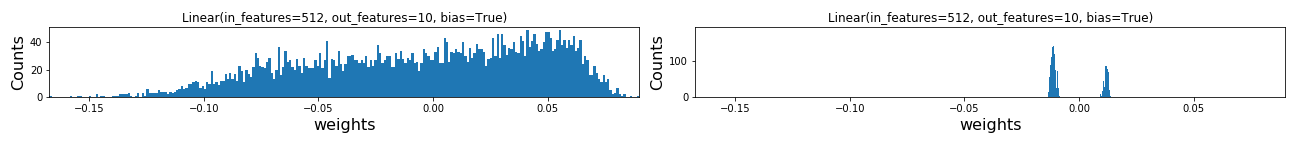}
    \includegraphics[width=0.8\textwidth]{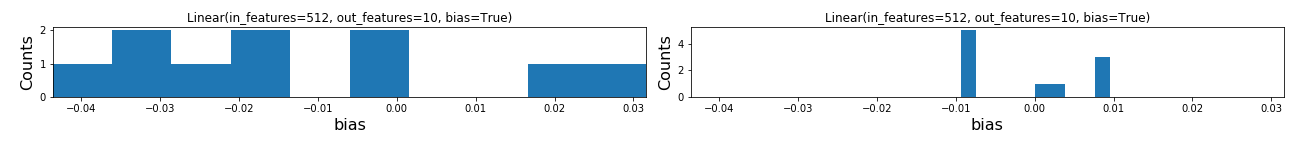}
    \vspace{-5.5mm}
    \caption{Weight distribution of ResNet18 trained on CIFAR10 with label noise ratio 0.4 (Cont'). Left: plain optimization; Right: volumized with $(v,\alpha)=(0.2, 0.99)$}
    \label{fig:weight-dist-0.4-2}
\end{figure}

\begin{figure}
    \centering
    \includegraphics[width=0.8\textwidth]{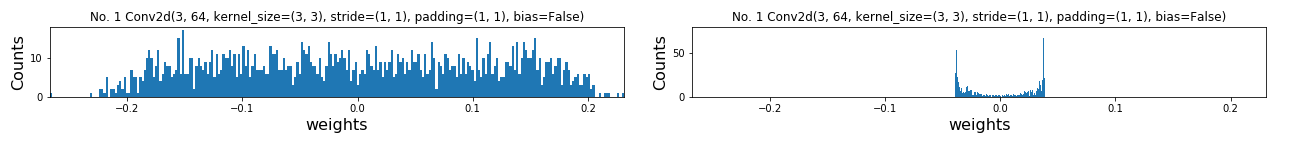}
    \includegraphics[width=0.8\textwidth]{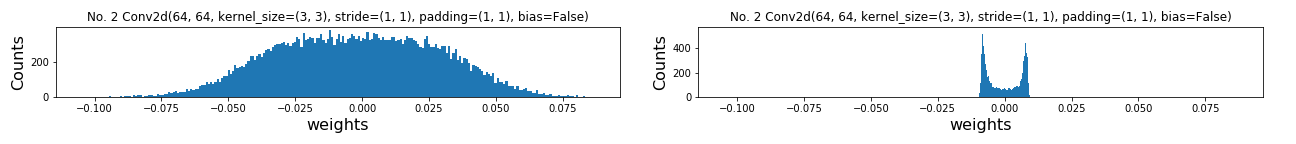}
    \includegraphics[width=0.8\textwidth]{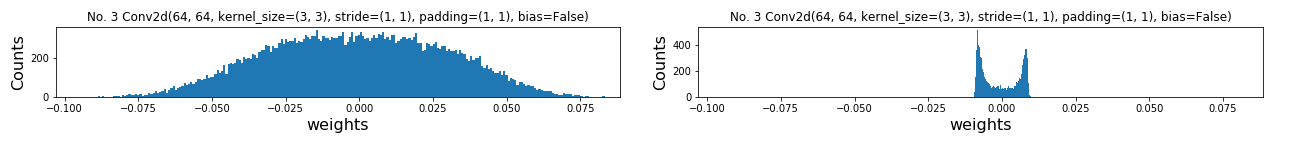}
    \includegraphics[width=0.8\textwidth]{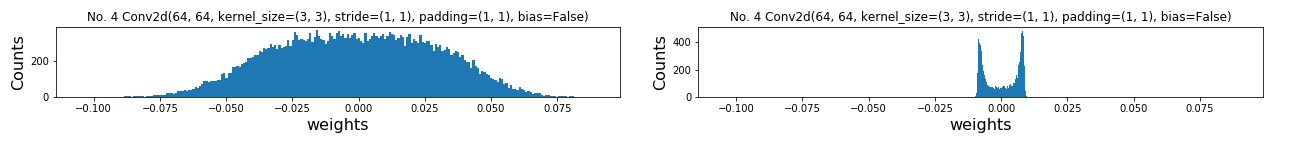}
    \includegraphics[width=0.8\textwidth]{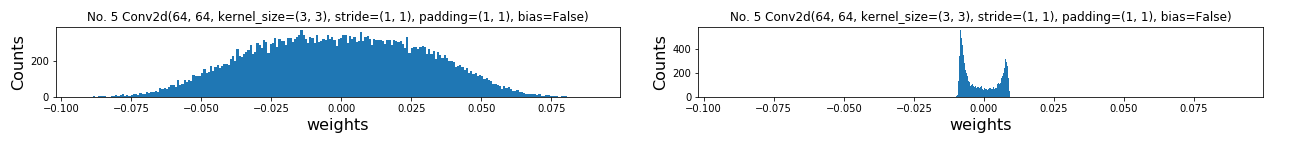}
    \includegraphics[width=0.8\textwidth]{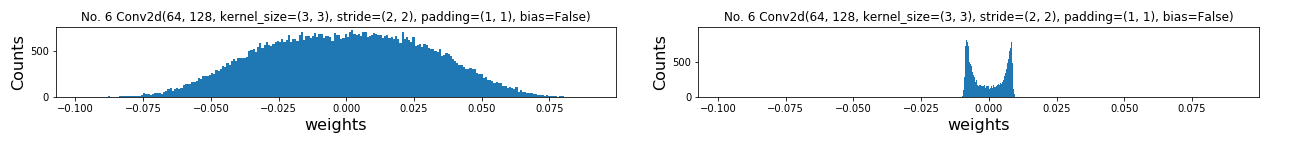}
    \includegraphics[width=0.8\textwidth]{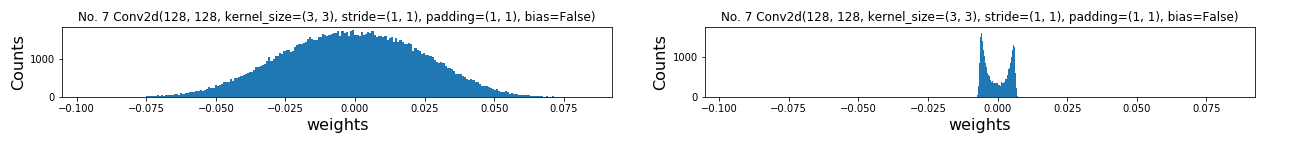}
    \includegraphics[width=0.8\textwidth]{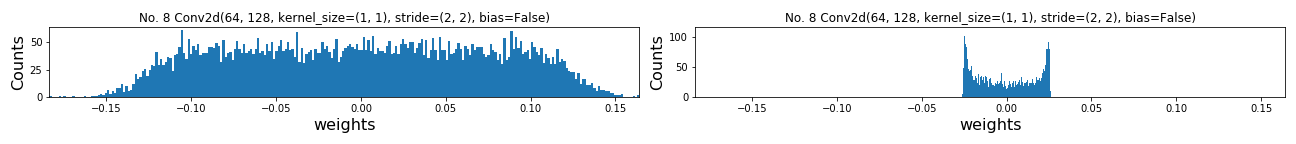}
    \includegraphics[width=0.8\textwidth]{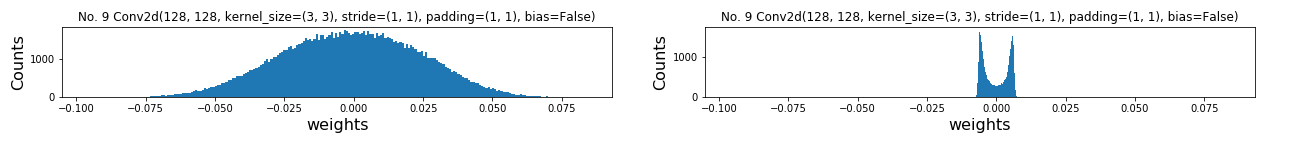}
    \includegraphics[width=0.8\textwidth]{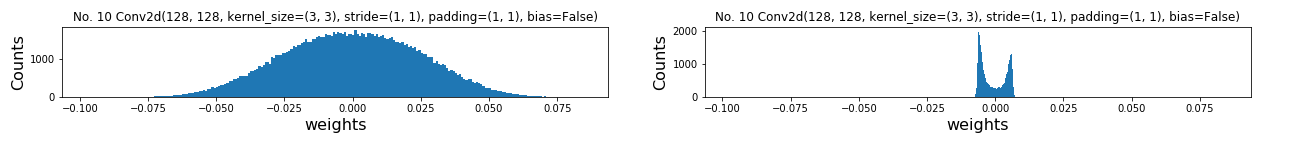}
    \includegraphics[width=0.8\textwidth]{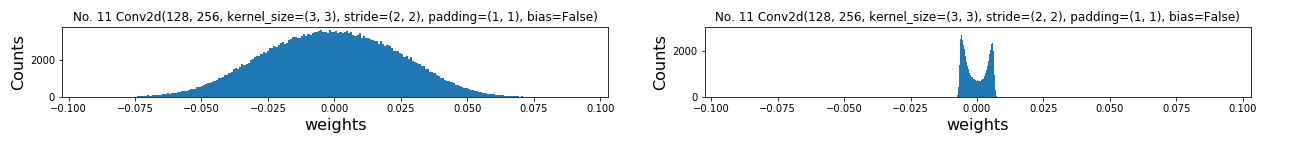}
    \includegraphics[width=0.8\textwidth]{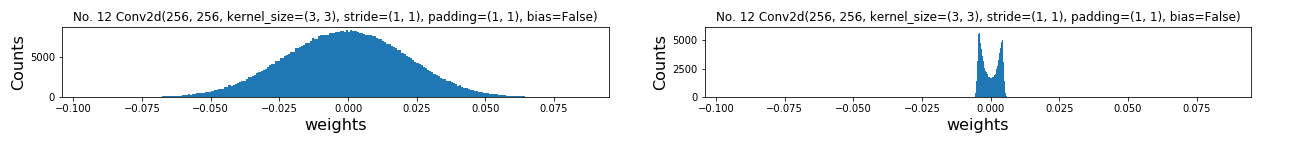}
    \includegraphics[width=0.8\textwidth]{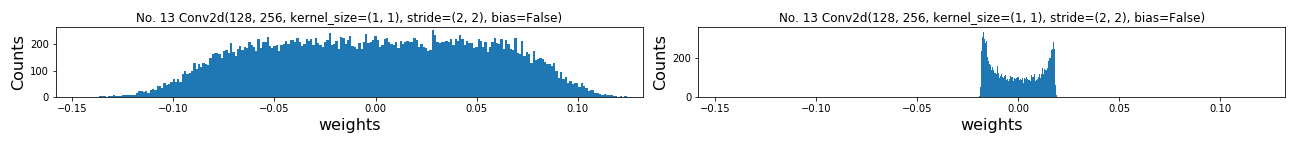}
    \includegraphics[width=0.8\textwidth]{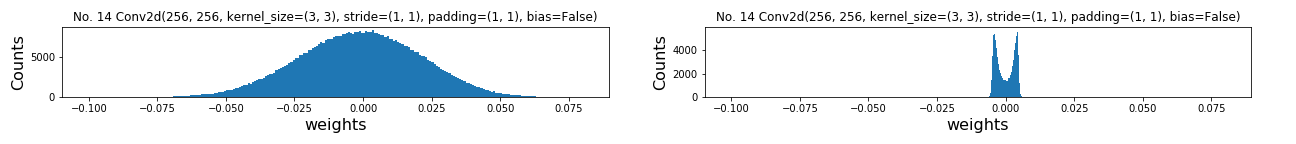}
    \vspace{-5.5mm}
    \caption{Weight distribution of ResNet18 trained on CIFAR10 with label noise ratio 0.8. Left: no volumization; Right: volumized with $(v,\alpha)=(0.2, 0.99)$}
    \label{fig:weight-dist-0.8-1}
    \vspace{1mm}
\end{figure}

\begin{figure}
    \centering
    \includegraphics[width=0.8\textwidth]{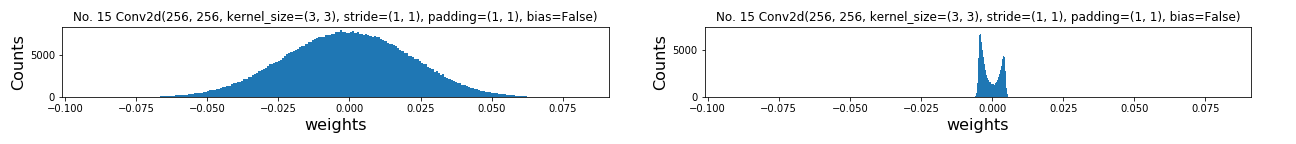}
    \includegraphics[width=0.8\textwidth]{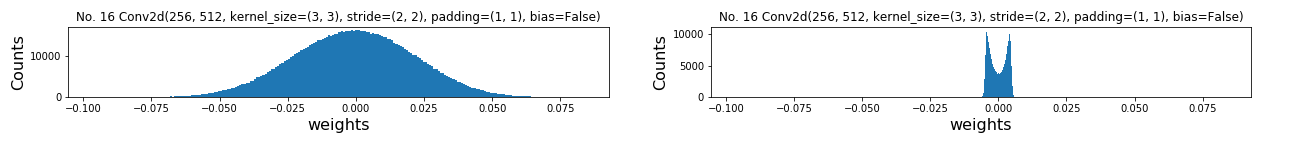}
    \includegraphics[width=0.8\textwidth]{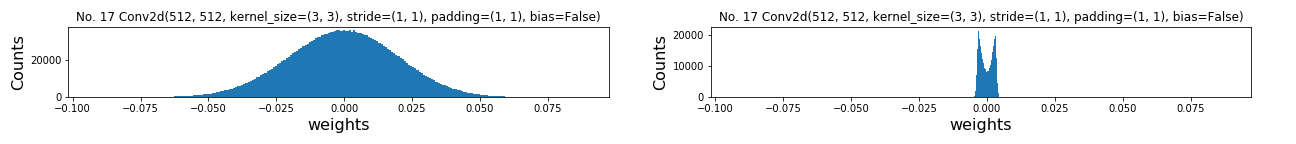}
    \includegraphics[width=0.8\textwidth]{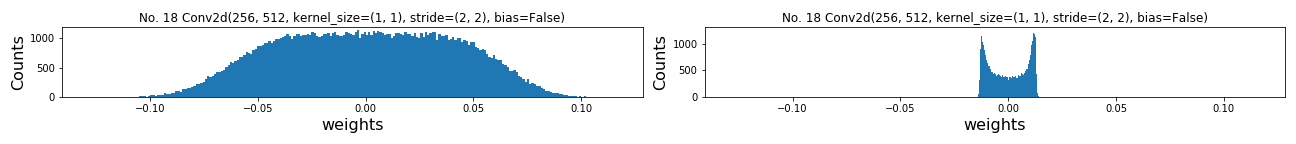}
    \includegraphics[width=0.8\textwidth]{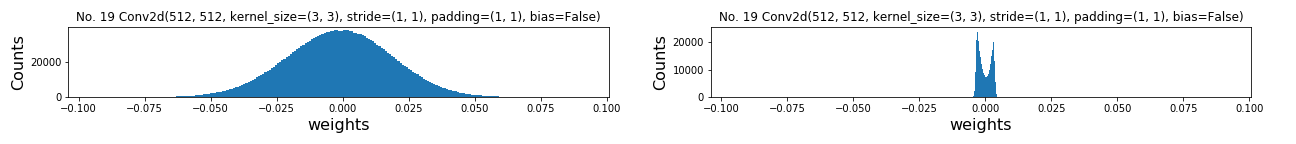}
    \includegraphics[width=0.8\textwidth]{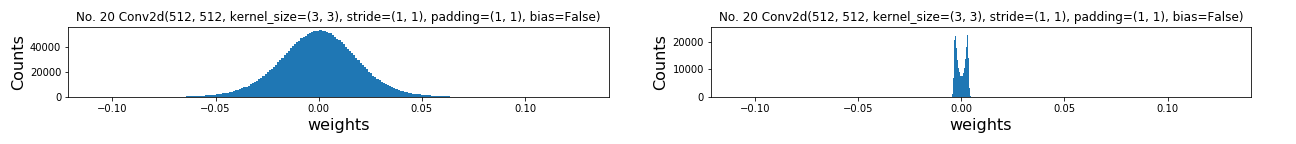}
    \includegraphics[width=0.8\textwidth]{plots/weight-dist/nr08-20conv.png}
    \includegraphics[width=0.8\textwidth]{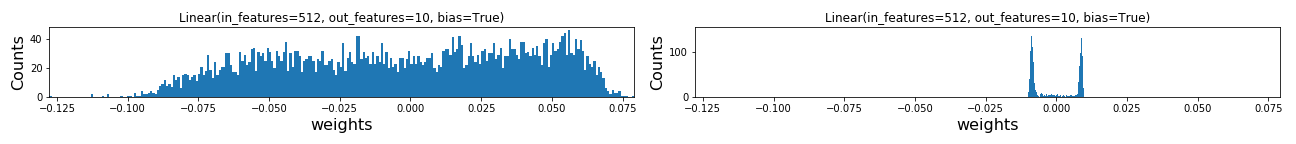}
    \includegraphics[width=0.8\textwidth]{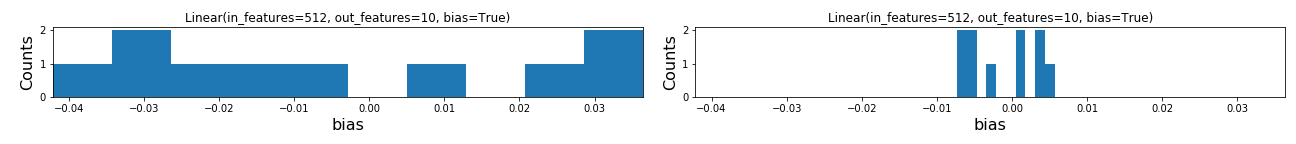}
    \vspace{-5.5mm}
    \caption{Weight distribution of ResNet18 trained on CIFAR10 with label noise ratio 0.8 (Cont'). Left: no volumization; Right: volumized with $(v,\alpha)=(0.2, 0.99)$}
    \label{fig:weight-dist-0.8-3}
\end{figure}

\vspace{-1mm}
In this section, we demonstrate further that volumization can be used to control the distribution of weight parameters in a neural network. In the main text, we demonstrated the peaking effect of a simple two-layer neural network, where the learned parameters are concentrated at $V$ and $-V$. Here, in Figure~\ref{fig:weight-dist-0.4-1} to Figure~\ref{fig:weight-dist-0.8-3}, we show that the same effect can be observed for a ResNet, for weights in all 20 convolutional layers and the last fully-connected layers as well as the bias in the last fully-connected layers. Notice that the volumized network also achieves better generalization performance according to the grid search in the previous section  This means that one can use volumization as a tool for training a neural network whose weights are binary (or ternary).

\end{document}